\documentclass[final,1p,times]{elsarticle}

\usepackage{amssymb}
\usepackage{amsmath}
\usepackage{float}
\usepackage{tikz}
\usepackage{upgreek}
\usepackage{eurosym}
\usepackage{bm}
\usepackage{tabularx} 
\usepackage{booktabs}
\usepackage{hhline}
\usepackage{url}

\begin{document}

\begin{frontmatter}



\title{Bayesian Neural Network Surrogates for Bayesian Optimization of Carbon Capture and Storage Operations}


\author[aff1]{Sofianos Panagiotis Fotias\corref{cor1}}
\ead{sfotias@metal.ntua.gr}         

\author[aff1,aff2]{Vassilis Gaganis}

\affiliation[aff1]{%
  organization={School of Mining and Metallurgical Engineering, National Technical University of Athens},
  postcode={157 73},
  city={Athens},
  country={Greece}
}
\affiliation[aff2]{%
  organization={Institute of Geoenergy, Foundation for Research and Technology},
  city={Chania},
  postcode={73100},
  country={Greece}
}

\cortext[cor1]{Corresponding author.}

\begin{abstract}
Carbon Capture and Storage (CCS) operations require optimizing injection and production controls under expensive, nonlinear reservoir simulations and operational constraints. Bayesian optimization (BO) is a practical derivative-free approach for this setting, traditionally relying on Gaussian process (GP) surrogates. This work benchmarks a range of stochastic surrogate models for BO in CCS operational optimization, including finite-width Bayesian neural networks (BNNs) with variational inference and Hamiltonian Monte Carlo variants, as well as alternative uncertainty-aware surrogates (MC dropout, deep ensembles, infinite-width BNNs, and deep kernel learning), against GP baselines. Two reservoir-simulation case studies are examined: (i) a low-dimensional single-objective control problem and (ii) a high-dimensional, multi-objective CCS scheduling problem with approximately 1,000 decision variables and an economic objective based on a simplified net present value (NPV). Across trials, BO produces substantial improvements over benchmark operations; however, surrogate rankings are problem-dependent and GPs remain difficult to outperform, even in the high-dimensional multi-objective setting. More exact Bayesian inference methods can incur markedly higher training cost without consistently improving BO outcomes. The results highlight the role of simulator-enforced constraints and nonlinear control realization in determining surrogate effectiveness and provide practical guidance for CCS BO model selection.
\end{abstract}

\begin{keyword}
CCS \sep Bayesian Optimization \sep Markov Chain Monte Carlo \sep Variational inference \sep Gaussian Process \sep Reservoir Simulation


\end{keyword}

\end{frontmatter}



\section{Introduction}
\label{sec1}
Achieving net-zero greenhouse gas emissions by mid-century necessitates significant investment in near-zero emissions technologies across industries \cite{jarvis2018technologies}. Carbon Capture, (Utilization) and Storage (CCUS or CCS) is a pivotal strategy in this transition, offering a flexible solution for decarbonizing diverse sectors by capturing CO$_2$ and sequestering it permanently in geological formations \cite{gabrielli2020role,bui2021role}. Viable storage sites, such as deep saline aquifers or depleted hydrocarbon reservoirs \cite{ji2015co2,bachu2015review,michael2010geological,hannis2017co2,mohammadian2019co,li2006co2}, must possess specific geological characteristics including adequate porosity, permeability and robust caprock integrity to ensure secure, long-term containment \cite{rackley2017introduction,tomic2018criteria}. Effective CCS deployment relies on meticulous site characterization and predictive numerical modeling to understand subsurface CO$_2$ behavior and trapping mechanisms \cite{ismail2023carbon,pruess2004code,class2009benchmark}.
\\\\
Numerical reservoir simulations are thus indispensable for evaluating CCS project performance as they simulate the fluid behavior through solving a discretized form of the Darcy equation \cite{whitaker1986flow}. In this study, OPM's open-source Flow software \cite{rasmussen2021open}, which employs a black-oil model formulation \cite{trangenstein1989mathematical} adapted for CO$_2$ brine systems, is utilized to simulate multi-phase compressible fluid flow and CO$_2$ trapping. Building on such accurate, albeit computationally intensive, simulation modeling, optimizing the development schedule is crucial for maximizing sequestration potential and economic viability in CCS projects \cite{cihan2015optimal,cameron2012optimization}. The primary goal of CCS operational optimization is to maximize CO$_2$ storage and/or economic returns, often expressed through the Net Present Value (NPV) function, while adhering to technical guidelines, operational constraints and commercial needs \cite{bachu2008co2,de2009acceptability}. Common strategies involve managing CO$_2$ injection rates and pressure buildup, often through brine production, with operational limits dictated by factors such as CO$_2$ breakthrough and maximum bottomhole pressures \cite{bandilla2017active,buscheck2016pre,anderson2020estimating,wang2012investigation}. Decision variables for CCS operators typically include well placement and time-varying fluid injection/production rates. This is challenging for long-horizon control schedules where the decision space becomes very high-dimensional and operational constraints can induce non-smooth objective responses. The resulting objective functions are often case-dependent, computationally expensive to evaluate (requiring full reservoir simulations) and exhibit complex, non-linear behavior, making them black-box functions suitable for multi-objective optimization \cite{deb2016multi}. While simulators internally handle some constraints (e.g., bottomhole pressures), others, such as preventing CO$_2$ migration beyond licensed areas or beyond structural traps, must be managed by the optimization process.
\\\\
The inherent complexity, non-linearity, local optima and numerous constraints in CCS optimization have led researchers to predominantly employ derivative-free optimization approaches \cite{islam2020holistic}. Methods such as Genetic Algorithms (GAs) \cite{mirjalili2019genetic, guyaguler2002optimization, el2006hybrid, emerick2009well} and Particle Swarm Optimization (PSO) \cite{kennedy1995particle, stopa2016optimization, hutahaean2014optimization}, often augmented with heuristics or global sampling strategies like Latin Hypercube Sampling (LHS) \cite{loh1996latin}, have been widely used for tasks like well placement and operational control in CCS and broader reservoir management \cite{badru2003well, goda2013global}. While effective for navigating non-smooth objective functions, these methods suffer from data inefficiency, particularly when direct objective function evaluations require computationally expensive, fully implicit reservoir simulations, as in this work.
\\\\
To mitigate this computational burden without resorting to potentially oversimplified proxy reservoir models, surrogate modeling has emerged as a preferred alternative \cite{cozad2014learning}. Surrogate models are computationally efficient approximations of the true objective function, built from a limited number of accurate simulation runs \cite{ong2003evolutionary}. Deterministic surrogates, such as radial basis functions, Kriging and multivariate adaptive regression splines, have been investigated for reservoir engineering optimization \cite{babaei2016performance, goel2007ensemble, santibanez2019maximising, a17100452}. However, careful selection of the surrogate strategy is crucial to avoid issues like convergence to non-existent optima \cite{jin2011surrogate}. Furthermore, deterministic models, especially those with many parameters like conventional neural networks, are prone to overfitting \cite{szegedy2013intriguing} and critically, cannot effectively quantify predictive uncertainty \cite{guo2017calibration,nixon2019measuring}. This lack of uncertainty quantification is particularly detrimental in data-scarce CCS optimization. Bayesian Optimization (BO) addresses these limitations by employing stochastic surrogate models \cite{frazier2018tutorial}. These models provide a robust mathematical framework to quantify both aleatoric uncertainty (e.g., from simulation noise or discrepancies between the model and reality) and epistemic uncertainty (stemming from a lack of data) \cite{der2009aleatory,hullermeier2021aleatoric}. BO iteratively constructs and updates these stochastic surrogates, most commonly Gaussian Processes (GPs) \cite{adler1990introduction}, to efficiently guide the search for the global optimum, making it well-suited for CCS optimization.
\\\\
BO, typically utilizing GP surrogates, has been successfully employed within reservoir engineering for diverse challenges. These include well placement optimization \cite{BALABAEVA202065}, field development planning under geological uncertainty to maximize NPV \cite{bordas2020bayesian}, optimizing gas/surfactant injection cycles in WAG schemes \cite{LU202296}, refining deviated well trajectories \cite{javed2020bayesian} and optimizing fracturing designs \cite{wang2017novel}. Innovative BO methodologies such as trust region BO and sparse-axis aligned subspaces BO (SAASBO) have also been proposed in the BO literature \cite{kumar2022search,kumar2023high,eriksson2019scalable,eriksson2021high,dou2015bayesian}. General advancements in BO research, extensively reviewed by Garnett et al. (2023) \cite{garnett2023bayesian}, have focused on areas like novel acquisition functions \cite{wang2017max}, expressive kernel designs \cite{gardner2017discovering,kandasamy2015high}, gradient-informed BO \cite{wu2017bayesian}, multi-objective optimization \cite{swersky2013multi} and extensions to discrete spaces \cite{moss2020boss}. Our own recent work contributed a permutation invariant kernel for CCS well placement optimization \cite{fotias2024optimization}. Despite these advancements and the successes of GP-based BO, controlled like-for-like comparative evidence on non-GP stochastic surrogate models under simulator-enforced constraints and long-horizon CCS control schedules remains limited.
\\\\
While Bayesian approaches, including Markov Chain Monte Carlo (MCMC) and specialized algorithms like NAB \cite{sambridge1999geophysical1,sambridge1999geophysical2}, are employed in reservoir engineering for tasks such as history matching and uncertainty propagation through development stages \cite{arnold2016optimisation, hutahaean2019reservoir, mohamed2010comparison, liao2019efficient, ma2008efficient, olalotiti2018multiobjective,buckle2019improving}, our focus here is different. This study investigates Bayesian Neural Networks (BNNs) \cite{neal2012bayesian} as alternative surrogate models to GPs within the BO framework, specifically for optimizing CCS operational decisions. MCMC techniques are utilized herein to sample from the BNN posterior over surrogate model parameters, not for history matching or uncertainty propagation across project phases.
\\\\
BNNs offer potential advantages over traditional GPs for complex optimization tasks. They can model non-stationary behavior and learn latent similarity measures via representation learning, features particularly valuable for high-dimensional inputs and multiple objectives common in CCS \cite{li2023study}. The recent development of Monte Carlo-based acquisition functions, which only require posterior samples from the surrogate, further facilitates the use of non-GP models like BNNs that may not have closed-form predictive distributions \cite{balandat2020botorch}. Therefore, this paper provides a controlled benchmark of BNN-family stochastic surrogates against GP baselines for Bayesian optimization of CCS operational controls, using a consistent BO pipeline adapted from Li et al. (2023) \cite{li2023study}. The primary contributions of this work are: (i) a comparative evaluation of multiple uncertainty-aware surrogate families under identical BO settings; (ii) assessment on both a low-dimensional control problem and a long-horizon, high-dimensional scheduling problem; and (iii) practitioner-relevant findings on performance trade-offs (sample efficiency versus training cost) and the impact of simulator-enforced constraints on surrogate behavior.
\\\\
The rest of the paper is organized as follows. Section \ref{sec2} provides an overview of the stochastic surrogate models implemented. Section \ref{sec3} details the CCS case studies used for optimization. Section \ref{sec4} presents and analyzes the optimization results. Section \ref{sec5} offers a broader discussion on the methodology, its application and performance. Finally, conclusions are drawn in Section \ref{sec6} based on the findings presented throughout the study.

\section{Methods}
\label{sec2}
\subsection{Gaussian Processes}
\label{subsec2.1}
Gaussian Processes (GPs) are the standard surrogate models in Bayesian Optimization (BO) due to their ease of implementation, data efficiency in low-sample regimes and inherent ability to quantify epistemic uncertainty through their kernel functions \cite{williams2006gaussian, frazier2018tutorial}. However, exact GP training scales cubically with the number of observations, motivating alternatives when the dataset grows. A GP defines a probability distribution over functions, characterized by a mean function, specifying the prior expected value at any input and a covariance (or kernel) function, quantifying the similarity and correlation between function values at different input points \cite{do2008multivariate, williams2006gaussian}. The kernel essentially encodes prior beliefs about the function's smoothness and behavior. In this work, the GP baseline was implemented using the default BoTorch SingleTaskGP configuration. An additional Appendix study compares this kernel with an explicit Matérn-5/2 GP to assess sensitivity to kernel choice. For comprehensive theoretical details and visual explanations, readers are referred to Rasmussen and Williams (2006) \cite{williams2006gaussian} and popular expositions \cite{agnihotri2020exploring, gortler2019visual}.
\\\\ 
Given a set of observations obtained by evaluating the expensive black-box function, the GP prior is updated to a posterior distribution over functions. This posterior provides a predictive mean and variance for any new potential candidate test point in the feasible space. These predictive distributions are then utilized by the acquisition functions (\ref{subsec2.4}) to decide where to sample next.
\subsection{Bayesian Neural Networks}
\label{subsec2.2}
While GPs are powerful, in practice BO often uses standard stationary kernels (e.g., Matérn or RBF/SE), which may be restrictive for highly non-stationary responses or high-dimensional inputs. BNNs offer a flexible alternative by placing prior distributions over the weights and biases of a neural network architecture \cite{goan2020bayesian, titterington2004bayesian, lampinen2001bayesian, jospin2022hands, li2023study}. Instead of learning point estimates for parameters as in standard ANNs, BNNs aim to infer a posterior distribution over these parameters given the observed data. This allows BNNs to inherently quantify predictive uncertainty and o: predictions from a BNN are obtained by marginalizing over this posterior distribution, effectively averaging over an ensemble of ANNs \cite{zhou2012ensemble, mackay1992practical}. However, the true posterior is generally intractable to compute directly due to the high-dimensional, non-convex nature of the parameter space and the difficulty in evaluating the evidence term \cite{izmailov2021bayesian}. Consequently, approximate inference techniques are required. The main approaches include Variational Inference (VI) and Markov Chain Monte Carlo (MCMC) sampling methods.
\\\\
\subsubsection{Variational Inference}
\label{subsec2.2.1}
Variational Inference (VI) approximates the intractable true posterior 
with a simpler, tractable distribution (e.g., mean-field Gaussians or more expressive normalizing-flow guides) \cite{blei2017variational}. The goal is to minimize the Kullback-Leibler (KL) divergence between the intractable posterior and the tractable one, which is equivalent to maximizing the Evidence Lower Bound (ELBO) \cite{kullback1951information} using stochastic gradient optimization. Stochastic Variational Inference (SVI) \cite{hoffman2013stochastic} is commonly used for this, with variational families ranging from mean-field (diagonal covariance) approximations to more expressive normalizing-flow guides \cite{graves2011practical, ritter2018scalable, maddox2019simple}. Further details on VI and ELBO maximization can be found in works like \cite{hernandez2015probabilistic, blundell2015weight, kingma2014adam, kingma2019introduction, kucukelbir2017automatic}. In this work, VI is implemented using Pyro’s SVI with an AutoIAFNormal guide and a Trace-ELBO objective optimized with Adam.

\subsubsection{Markov Chain Monte Carlo}
\label{subsec2.2.2}
As an alternative to approximating the posterior directly, sampling methods aim to draw samples from the BNN posterior. These samples are then used to estimate the predictive distribution for BO \cite{hastings1970monte, gilks1995markov}. While basic samplers like random-walk Metropolis or slice sampling \cite{neal2003slice} have limitations in high dimensions, MCMC algorithms such as Metropolis-Hastings construct a Markov chain whose stationary distribution is the target posterior \cite{bardenet2017markov, norris1998markov, chib1995understanding}.
\\\\
For complex, high-dimensional posteriors common in BNNs, Hamiltonian Monte Carlo (HMC) and its variants, like the No-U-Turn Sampler (NUTS) \cite{hoffman2014no}, are often more efficient \cite{neal2012mcmc, betancourt2017conceptual}. HMC introduces auxiliary momentum variables and leverages Hamiltonian dynamics to propose distant, high-acceptance-probability samples, effectively navigating the geometry of the typical set of the distribution. This is particularly advantageous in exploring regions of high curvature where simpler random-walk MCMC methods struggle. HMC methods typically require gradient information of the log-posterior and careful tuning of parameters, often involving symplectic integrators to maintain numerical stability \cite{girolami2011riemann, leimkuhler1994symplectic, yoshida1993recent}. In this work, we use Pyro’s NUTS implementation (with JIT compilation enabled) to draw posterior samples of the BNN parameters.

\subsection{Approximate Bayesian neural network surrogates}
\label{subsec2.3}
Beyond VI and MCMC, several computationally less intensive 'approximate Bayesian' techniques are used to obtain uncertainty estimates from neural networks, making them suitable for BO surrogates.
\subsubsection{MC dropout}
\label{subsubsec2.3.1}
Dropout \cite{JMLR:v15:srivastava14a} is a method initially developed for regularization of deep networks. The key idea is to randomly drop units from each layer during training thereby preventing units from co-adapting. MC dropout \cite{pmlr-v48-gal16} is a method that builds on this idea by keeping the dropout probabilities when making predictions on new inputs. For each input, multiple stochastic forward passes are performed and a mean and variance are estimated from these samples. MC Dropout has been shown to approximate a specific form of VI under certain conditions \cite{pmlr-v48-gal16}. This method is very convenient and straightforward to implement.
\subsubsection{Deep Ensembles}
\label{subsubsec2.3.2}
Deep Ensembles \cite{lakshminarayanan2017simple} is another straightforward and easily implemented method. This method trains multiple neural networks on the same training data (optionally with bootstrap resampling), using different random initializations. The posterior is approximated with a distribution parameterized as the sum of multiple Dirac delta functions. These methods provide a good predictive uncertainty estimate \cite{ovadia2019can}. 

\subsubsection{Infinite width Neural Networks}
\label{subsubsec2.3.3}
At initialization, a deep neural network in the limit of infinite width was shown to induce a GP prior \cite{neal1996priors,williams1996computing} using the Central Limit Theorem. This can be extended for not only a single network but for any number of additional hidden layers as well \cite{lee2017deep}. The kernel function of this GP is complex and can be derived from the architecture of the network and its computation is deterministic and differentiable.

\subsubsection{Deep Kernel Learning}
\label{subsubsec2.3.4}
Deep Kernel Learning (DKL) \cite{wilson2016deep} is a hybrid method where a neural network transforms the input features into a new representation, upon which a standard GP kernel is then applied. This allows the model to learn complex, non-stationary feature mappings before leveraging the probabilistic framework of GPs.

\subsection{Acquisition functions}
\label{subsec2.4}
The BO loop iteratively selects new points for evaluation using an acquisition function, which leverages the surrogate model's predictive mean and variance to balance exploration (sampling in uncertain regions) and exploitation (sampling near known good values). For single-objective optimization, a common choice is Expected Improvement (EI). EI quantifies the expected amount by which a new candidate point will improve upon the best objective value observed so far. When a closed-form posterior from the surrogate is unavailable (as with BNNs), a Monte Carlo approximation (MC-EI) is used where posterior samples of the surrogate model parameters are drawn, predictions are made at candidate points and the expected improvement is averaged over these samples. The point maximizing this MC-EI is chosen for the next evaluation .
\\\\
For multi-objective optimization problems, as encountered in this study, EI is extended to Expected Hypervolume Improvement (EHVI). EHVI measures the expected increase in the hypervolume of the objective space dominated by the current Pareto front of non-dominated solutions, given a new set of candidate points. Similar to MC-EI, Monte Carlo EHVI (MC-EHVI) is computed by averaging the hypervolume improvement over predictions from posterior samples of the surrogate model. The batch of candidate points that maximizes MC-EHVI is then selected for evaluation. This approach guides the search towards solutions that broadly improve across all objectives by expanding the dominated region of the Pareto front.
\\\\
In this implementation, single-objective optimization was performed using Monte Carlo q-Expected Improvement (qEI), while multi-objective optimization was performed using Monte Carlo q-Expected Hypervolume Improvement (qEHVI). Batch candidate generation was performed by direct joint optimization of the corresponding q-acquisition function in the $q\times d$-dimensional candidate space, rather than by sequential selection. In the discrete CCS problems, optimized candidates were then mapped back to the original domain, rounded to feasible integer-valued controls, and rescored after rounding before the final batch was selected for simulator evaluation. For consistency across all surrogate families, the Monte Carlo formulations were also retained for the GP-based experiments rather than using analytic EI or analytic EHVI specifically for GPs. The Monte Carlo approximation was based on 2048 posterior samples. The acquisition strategy was intentionally kept fixed across surrogate models. This was done to isolate the effect of surrogate choice within a common BO framework and avoid introducing acquisition-function choice as an additional experimental variable. A broader comparison with alternative acquisition functions is left for future work.

\section{Problem description}
\label{sec3}
\subsection{The reservoir system}
\label{subsec41}
\begin{figure}[H]
\includegraphics[width=0.8\linewidth,keepaspectratio]{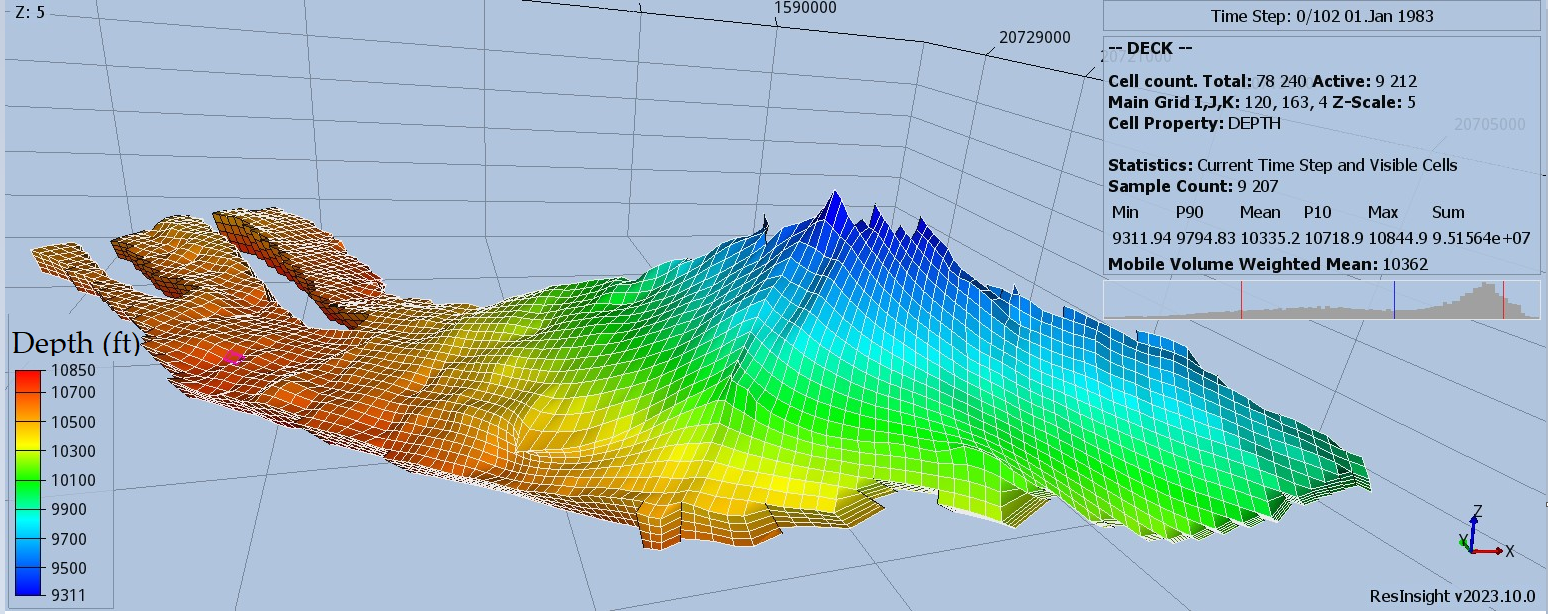}
\caption{Aquifer depth (z axis is scaled) \label{fig2}}
\end{figure}

In this study, we reuse the synthetic aquifer benchmark from our previous work \cite{fotias2024optimization}, which examined the carbon sequestration potential of a deep, highly permeable synthetic aquifer. As illustrated in Figure \ref{fig2} (visualized in ResInsight), the aquifer exhibits an inclined geometry reminiscent of an anticline. It spans a large horizontal area and is composed of four sedimentary layers, each characterized by different average permeability values. The formation is tightly sealed by shale; no faults or fractures are included in the model and salinity is approximately 65,000 ppm. The temperature profile of the aquifer adheres to expected thermal gradients at its depth, as detailed in Table \ref{tab1}. Due to its high permeability, the pressure distribution within the aquifer aligns with the pressure gradient and remains isotropic in the $x$-$y$ plane. However, the aquifer is slightly overpressurized, which necessitates the simultaneous extraction of brine during CCS operations. This approach is critical for postponing the attainment of the upper pressure safety limit of 9,000 psi mandated by technical and safety protocols. The significance of this strategy is further underscored by the no-flow boundary conditions imposed at the aquifer's lateral bounds in the simulation model \cite{peaceman2000fundamentals}.

\begin{figure}[H]
\includegraphics[width=0.8\linewidth,keepaspectratio]{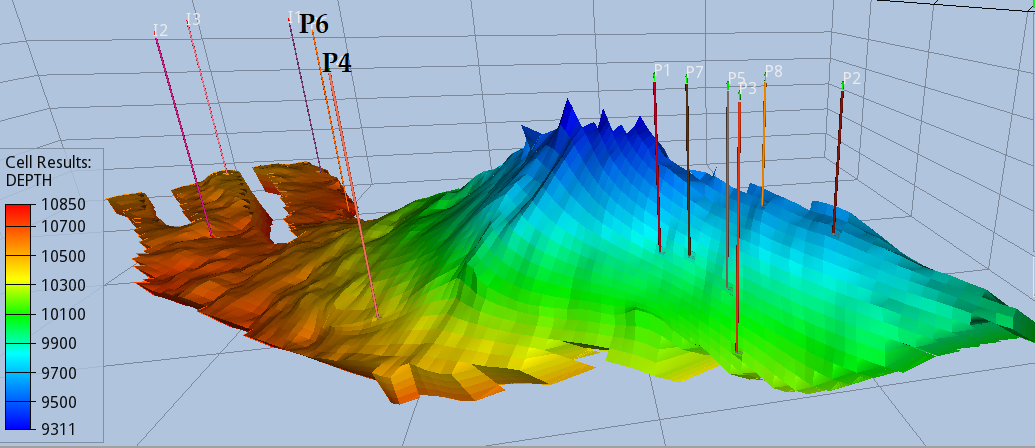}
\caption{Optimal well placements \label{fig3}}
\end{figure}

\begin{figure}[H]
\includegraphics[width=0.7\linewidth,keepaspectratio]{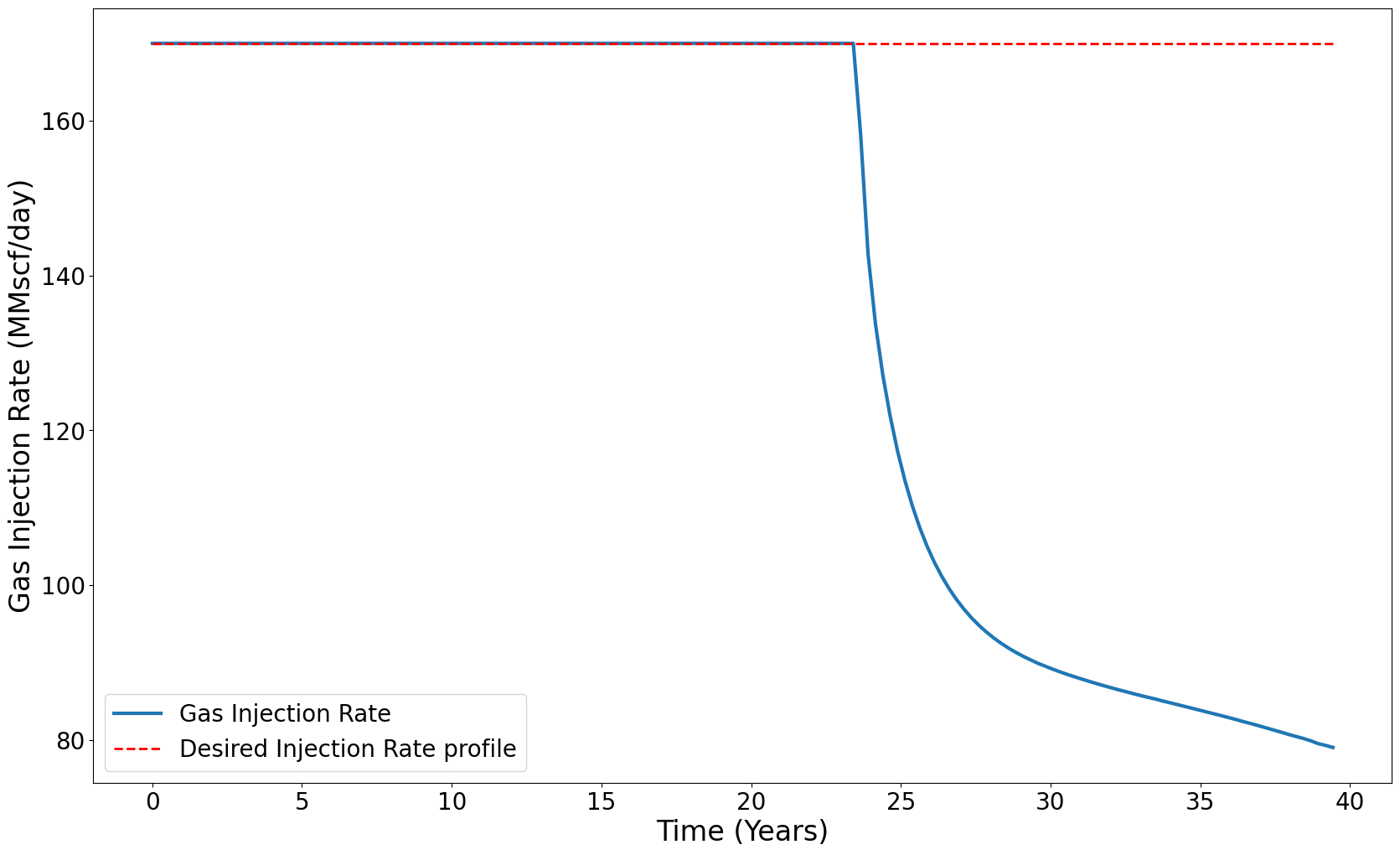}
\caption{Placement optimum configuration injection performance \label{fig4}}
\end{figure}

\begin{table}[H] 
\caption{Aquifer static properties (average values)\label{tab1}}
\newcolumntype{C}{>{\centering\arraybackslash}X}
\begin{tabularx}{\textwidth}{CCC}
\toprule
\textbf{Parameter}	& \textbf{Value}	& \textbf{Units}\\
\midrule
Pressure ($P$) & $5,000$ &$Psi$\\
Temperature ($T$) &$200$ &$^\circ F$\\
Porosity ($\phi$) & 0.25 &\\
Depth ($D$) & $10,180$ & $ft$\\
Permeability ($k$)		& $300$ 			& $mD$\\
Bulk volume ($V$)	& 		$2.4\cdot10^{11}$	& $scf$\\
Water in place & $1.7\cdot10^{9}$ & $STB$\\
\bottomrule
\end{tabularx}
\end{table} 
In our previous work, BO was used with a GP surrogate model to optimize well placement of three injectors and eight producers. Exploiting the symmetry of injector and producer locations under the group control policies, a permutation invariant kernel was developed that rapidly expedited the optimization. The objective function selected was the total CO$_2$ sequestrated mass based on predefined target injection and production rates. The optimal configuration is seen in Figure \ref{fig3}.
All injectors are placed at the bottom of the reservoir, on the left side of Figure \ref{fig3}. A cluster of six producers has been strategically placed at the opposite side of the aquifer as far away as possible from the injectors, with the exception of two producers, highlighted in the figure, placed closer to the injectors for pressure maintenance. This specific placement and the predefined scheduling will be referred to throughout this work as the "placement optimum" configuration for easy reference. Note that the producer count was chosen to highlight the benefit of the permutation-invariant kernel in the well-placement benchmark \cite{fotias2024optimization}. 
\\\\
Building upon this, the attention now turns to optimizing the injection and production rates over the envisaged injection period. Based on the results of the optimal configuration achieved, the CO$_2$ injection rate profile can be seen in the blue line of Figure \ref{fig4}. In this work, two approaches will be utilized for optimization, one with a single objective function and a few decision variables ($\approx 10$) and a multi-objective optimization problem with many decision variables ($\approx 1000$). The finite-width BNN surrogates (SVI, NUTS, Dropout) use three hidden layers of 100 fully connected neurons each. The models are named as seen in Table \ref{tab3}.
\begin{table}[H]
\caption{Model names\label{tab3}}
\centering
\begin{tabular}{|>{\raggedright}p{3cm}|>{\raggedright\arraybackslash}p{10cm}|}
\hline
Model Name & Model Description \\
\hline
GP & Gaussian Process with RBF/SE kernel (Section \ref{subsec2.1})\\
\hline
SVI & Stochastic Variational Inference (Section \ref{subsec2.2.1}) \\
\hline
MCMC & Hamiltonian Monte Carlo (Section \ref{subsec2.2.2}) \\
\hline
NUTS & Hamiltonian Monte Carlo with No-U-Turn Sampler (Section \ref{subsec2.2.2}) \\
\hline
IBNN & Infinite Width Bayesian Neural Network (Section \ref{subsubsec2.3.3}) \\
\hline
Ensemble & Deep Ensembles (Section \ref{subsubsec2.3.2}) \\
\hline
Dropout & MC dropout (Section \ref{subsubsec2.3.1}) \\
\hline
DKL & Deep Kernel Learning (Section \ref{subsubsec2.3.4}) \\
\hline
\end{tabular}
\end{table}
The BO framework is built upon Li's work \cite{li2023study} that utilizes Gpytorch \cite{gardner2018gpytorch} and Botorch \cite{balandat2020botorch} for the models and BO with the addition of Pyro \cite{bingham2019pyro} for the NUTS and SVI models. The framework consists of selecting 15 initial points for evaluation and subsequently performing 15 BO iterations for each model, selecting the 4 best candidate points to be evaluated and updating the training dataset each time. This results in 15 initial simulations plus 15 batches of 4 evaluations (75 simulations per model/run).

\subsection{Case study 1}

In the placement optimum case, while the target injection rate of 170 MMscf/day remains constant for $\sim23$ years and then drops rapidly. Due to gas breakthrough, producers operate below their target rates, resulting in a pressure increase and a subsequent drop in gas injection rate due to the max bottomhole pressure constraint. In this case study, the group control policy of the injection wells is necessary since ideally, a constant total sequestration rate would be essential to honor the operator's contracts to the emitters, therefore it needs to remain constant and it has been kept at the same value of 170 MMscf/day. However, in search for more operational decision variables and increased flexibility, the group control policy has been removed from the producers. By making them independent, BO is used to identify target production rates for each well independently so as the target injection rate is honored throughout the project's lifespan without unnecessary CO$_2$ recycling. This is a classic single objective optimization problem, formulated in Eq. \ref{eq28}.
\begin{equation}
    \label{eq28}
    \begin{aligned}
    & \underset{\mathbf{x}}{\text{maximize}}
& & f(\mathbf{x}) = 1/\left[\sum_t \left|Q_{CO_2}^{\text{inj}}(t) - Q_{CO_2}^{\text{inj}}(t-1)-Q_{CO_2}^{\text{prod}}(t)\right|\cdot k(t)\right] \\
& \text{subject to}
& & g_i(\mathbf{x}) \leq 0, \quad i = 1,2,\ldots,m \\
\end{aligned}
\end{equation}
where $f(\mathbf{x})$ penalizes deviations from constant injection and also penalizes CO$_2$ recycling per timestep through the term $Q_{CO_2}^{\text{prod}}(t)$. The term $k(t)\propto 1/t$ is utilized to penalize harder the injection rate drop at the initial stages of the project as opposed to the latter ones. This objective is sampled by simulation runs. Furthermore $\mathbf{x}$ are the decision variables, in this case the target production rate of each producer and the breakthrough control. The constraints $g_i(\mathbf{x}) \leq 0$, are the bounds of the decision variables and the bottomhole pressure constraints which are handled by the simulator and thus need not be introduced directly in the optimization formulation. Decision variables and constraints can be seen in detail in Table \ref{tab4}
\begin{table}[H] 
\caption{Decision variables for first variation of Case study 1\label{tab4}}
\newcolumntype{C}{>{\centering\arraybackslash}X}
\begin{tabularx}{\textwidth}{lCC}
\toprule
\textbf{\# variables}&\textbf{Variable type} & \textbf{Range} \\
\midrule
 8&(Target) Production Rate (Water) & $\leq 100$ Mstb/day \\
8&(Constraint) Max Production Rate (CO$_2$) & $\leq 8$ MMscf/day\\
\bottomrule
\end{tabularx}
\end{table}
Since in practice, BO managed to easily converge to an optimal scenario where injection rate is kept constant throughout the development, a second variation of this case is also examined by trying to increase the injection rate, using it as an extra decision variable rather than fixing it at 170 MMscf/day. The decision variables of the second variation of Case study 1 can be seen in Table \ref{tab5}.
\begin{table}[H] 
\caption{Decision variables for second variation of Case study 1 \label{tab5}}
\newcolumntype{C}{>{\centering\arraybackslash}X}
\begin{tabularx}{\textwidth}{lCC}
\toprule
\textbf{\# variables}&\textbf{Variable type} & \textbf{Range} \\
\midrule
1& (Target) Injection Rate (CO$_2$) & $170-200$ MMscf/day \\
 8&(Target) Production Rate (Water) & $\leq 100$ Mstb/day \\
8&(Constraint) Max Production Rate (CO$_2$) & $\leq 8$ MMscf/day \\
\bottomrule
\end{tabularx}
\end{table}
This variation was handled as a multi-objective optimization case with the first objective being the same as in the first variation and the second being the total sequestration, for the optimizer to be guided towards larger injection rates.
\begin{equation*}
    f_2(\mathbf{x}) =  \sum_t \left(Q_{CO_2}^{\text{inj}}(t) - Q_{CO_2}^{\text{prod}}(t)\right)
\end{equation*}
Although a target rate limit of 100 Mstb/day for a producer might be way over the maximum capacity of what a well can handle, reservoir simulators enforce physical constraints (such as bottomhole pressure limits), so any target rate exceeding feasible bounds defaults to the same practical outcome. In other words, infeasibly high targets are clipped by simulator constraints; nevertheless, different optimizers converge to distinct feasible solutions. 

\subsection{Case study 2}
\begin{figure}[H]
\includegraphics[width=0.7\linewidth,keepaspectratio]{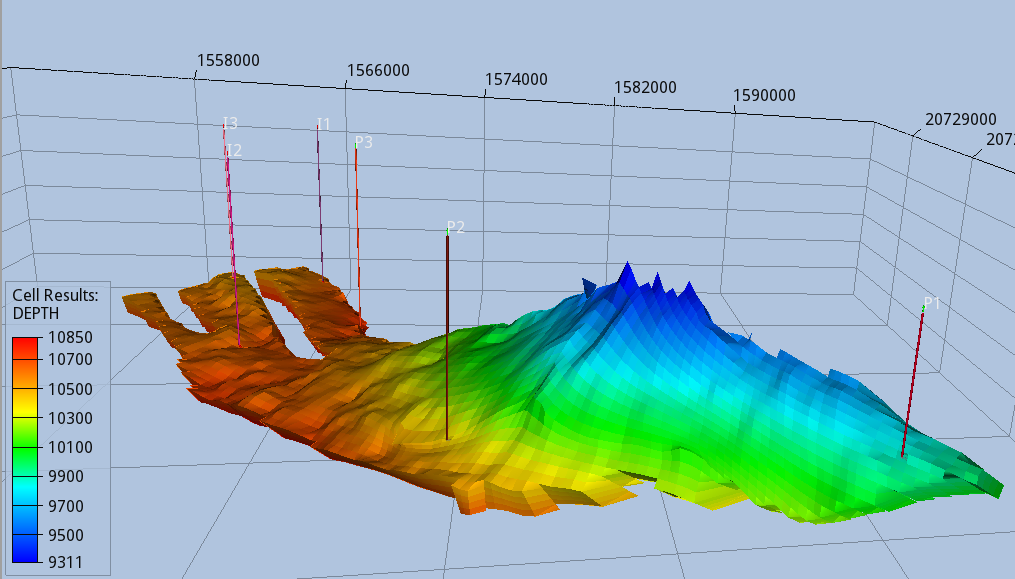}
\caption{reduced placement optimum \label{fig6}}
\end{figure}

\begin{figure}[H]
\includegraphics[width=0.7\linewidth,keepaspectratio]{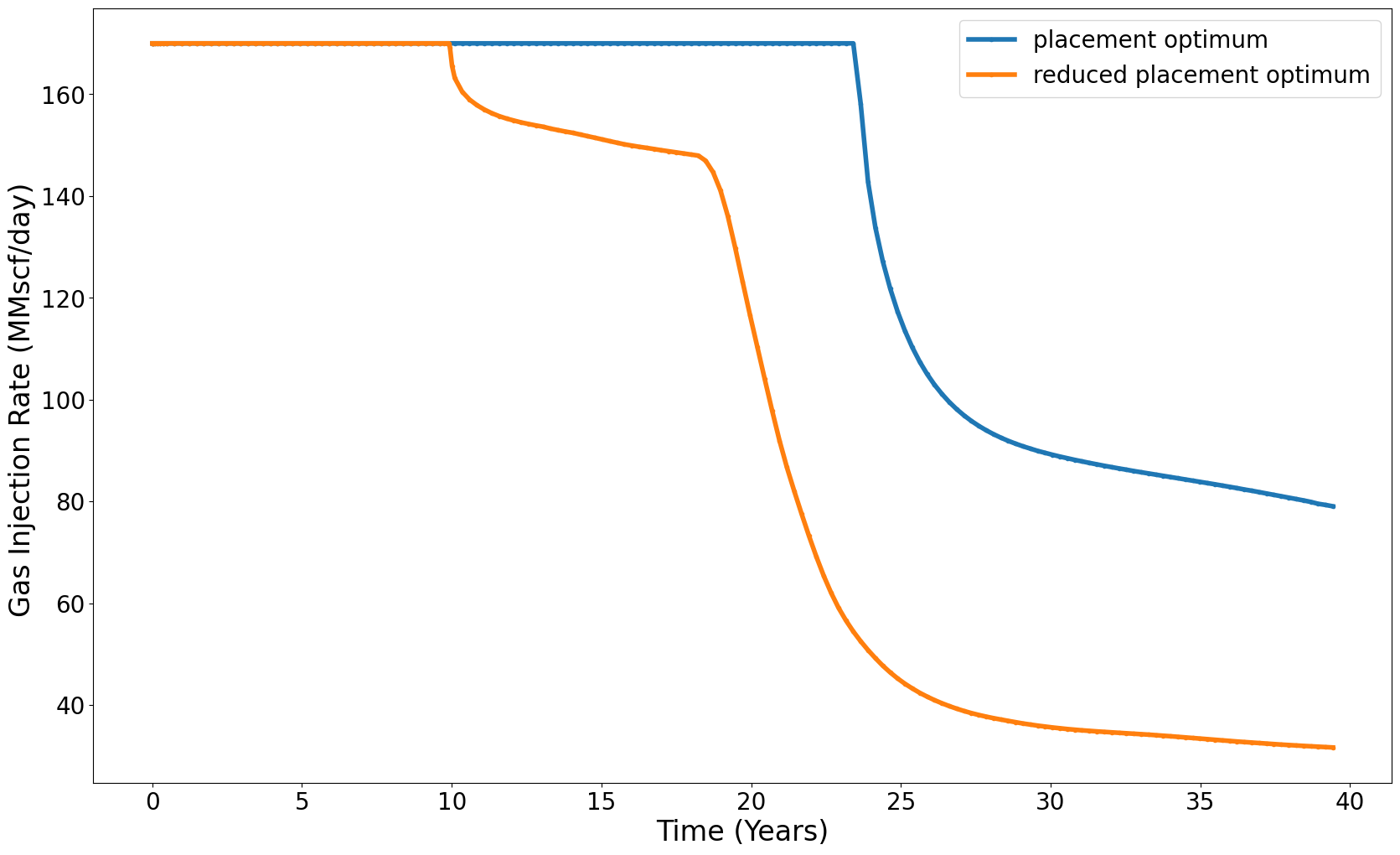}
\caption{Gas injection rate profile \label{fig5}}
\end{figure}

In the second case, the cluster of six producers in the placement-optimum configuration is replaced by a single producer, while the remaining two producers are retained, resulting in a reduced configuration with three producers in total (Figure \ref{fig6}), to emulate operations capital expenditure (CAPEX) reduction goals.
Although it is not typical to budget the drilling of 6 wells for pressure maintenance as each one's cost is in the range of tens of millions of dollars, it needs to be noted that most producers in the cluster could function as lateral extensions of a few multi-segment horizontal wells thereby reducing the costs. Regardless, by performing this aggregation, the total amount of CO$_2$ sequestered drops significantly as seen in Figure \ref{fig5}. This case will be referred to as the "reduced placement optimum" one.

Similar to the previous case study, the group control policy of the three producers retained is removed. Their target rate as well as their allowable breakthrough rate is adjusted by the optimizer every 90 days. We discretize the 40-year horizon into 160 control intervals (90 days each), yielding $3\times160 = 480$ rate variables for the three producers (and the same number for CO$_2$ limits). Furthermore, a single decision variable is used for the target injection rate, slightly tunable from the base value of 170 MMscf/day. The decision variables are shown in Table \ref{tab6}.
\begin{table}[H] 
\caption{Decision variables for Case study 2 \label{tab6}}
\newcolumntype{C}{>{\centering\arraybackslash}X}
\begin{tabularx}{\textwidth}{lCC}
\toprule
\textbf{\# variables}&\textbf{Variable type} & \textbf{Range} \\
\midrule
1& (Target) Injection Rate (CO$_2$) & $150-190$ MMscf/day \\
 480&(Target) Production Rate (Water) & $\leq 100$ Mstb/day \\
480&(Constraint) Max Production Rate (CO$_2$) & $\leq 8$ MMscf/day\\
\bottomrule
\end{tabularx}
\end{table}

In this optimization case, four objective functions were utilized. The goal is to tune the reduced placement optimum controls to recover, as much as possible, the placement optimum performance across four CCS-relevant objectives. The four objective functions are shown below in Eq. \ref{eq:objectives}  

\begin{equation}
    \label{eq:objectives}
    \begin{aligned}
    & \underset{\mathbf{x}}{\text{maximize}}
    & & f_1(\mathbf{x}) = \frac{M_{CO_2}^{\text{residual}} + M_{CO_2}^{\text{dissolved}}}{M_{CO_2}^{\text{mobile}}} \\
    & \underset{\mathbf{x}}{\text{maximize}}
    & & f_2(\mathbf{x}) =  \sum_t \left(Q_{CO_2}^{\text{inj}}(t) - Q_{CO_2}^{\text{prod}}(t)\right)\\
    & \underset{\mathbf{x}}{\text{maximize}}
    & & f_3(\mathbf{x}) = 1/\left[\sum_t \left|Q_{CO_2}^{\text{inj}}(t) - Q_{CO_2}^{\text{inj}}(t-1)+1\right|\cdot k(t) \right]\quad  \\
    & \underset{\mathbf{x}}{\text{maximize}}
    & & f_4(\mathbf{x}) = \sum_t \Delta t_n \left( C_{\text{storage}} \cdot Q_{CO_2}^{\text{stored}} - C_{\text{injection}} \cdot Q_{CO_2}^{\text{inj}}- C_{\text{production}} \cdot Q_{\text{brine}}\right)/(1+r)^{t/365} \\
    & \text{subject to}
    & & g_i(\mathbf{x}) \leq 0, \quad i = 1,2,\ldots,m
    \end{aligned}
\end{equation}
The first objective aims at maximizing the desired trapping mechanisms like solubility of CO$_2$ in brine and residual which is calculated based on the amount of immovable CO$_2$ left as the plume moves through the aquifer. The second one is the same as the one used in the second variation of the first case study. The third is the same as the first case study's first variation, without the term penalizing CO$_2$ recycling as this has now been indirectly introduced to the fourth objective which represents the project Net Present Value (NPV). Approximating the NPV of a CCS project involves a lot of uncertainty since there are not many in operation worldwide yet. Various simplifications have been made in order to use this functional form of the NPV. Costs associated with recasing wells or even drilling new ones have not been dealt with despite potentially being substantial \cite{platform2011costs}. The cost of injecting CO$_2$ $C_{\text{injection}}$ has thus been placed at a moderate 6.2 \euro \;per tonne injected. The profit of storing emissions $C_{\text{storage}}$ is estimated at around 30 \euro \;per tonne of CO$_2$. This is an indicative expected value defined by IEA \cite{iea2019world} but it is based on the assumption that it will be a closed contract between the emitter and the CCS operator and not be purchased on the spot market. In fact, due to the fast changing environment of regulations regarding CO$_2$ emissions, there is a lot of uncertainty surrounding this price in general. Finally the cost of producing and treating brine $C_{\text{production}}$ was set at 2.7 \euro \;per tonne \cite{santibanez2019maximising,sullivan2013method}. For this objective, CO$_2$ rates/volumes are converted to mass units (tonnes) using consistent surface-condition assumptions. The annual discount rate of the NPV was set at $r = 1.42\%$. This objective function is not meant to provide an exact measure of the profitability of the CCS project. It doesn't account for inflation, initial investment and operational expenses, it is rather meant to serve as a counterweight to extreme brine production and CO$_2$ breakthrough which hinders profitability.

\section{Results}
\label{sec4}
\subsection{Case study 1}
\subsubsection{Variation 1}
\begin{figure}[H]
\includegraphics[width=0.7\linewidth,keepaspectratio]{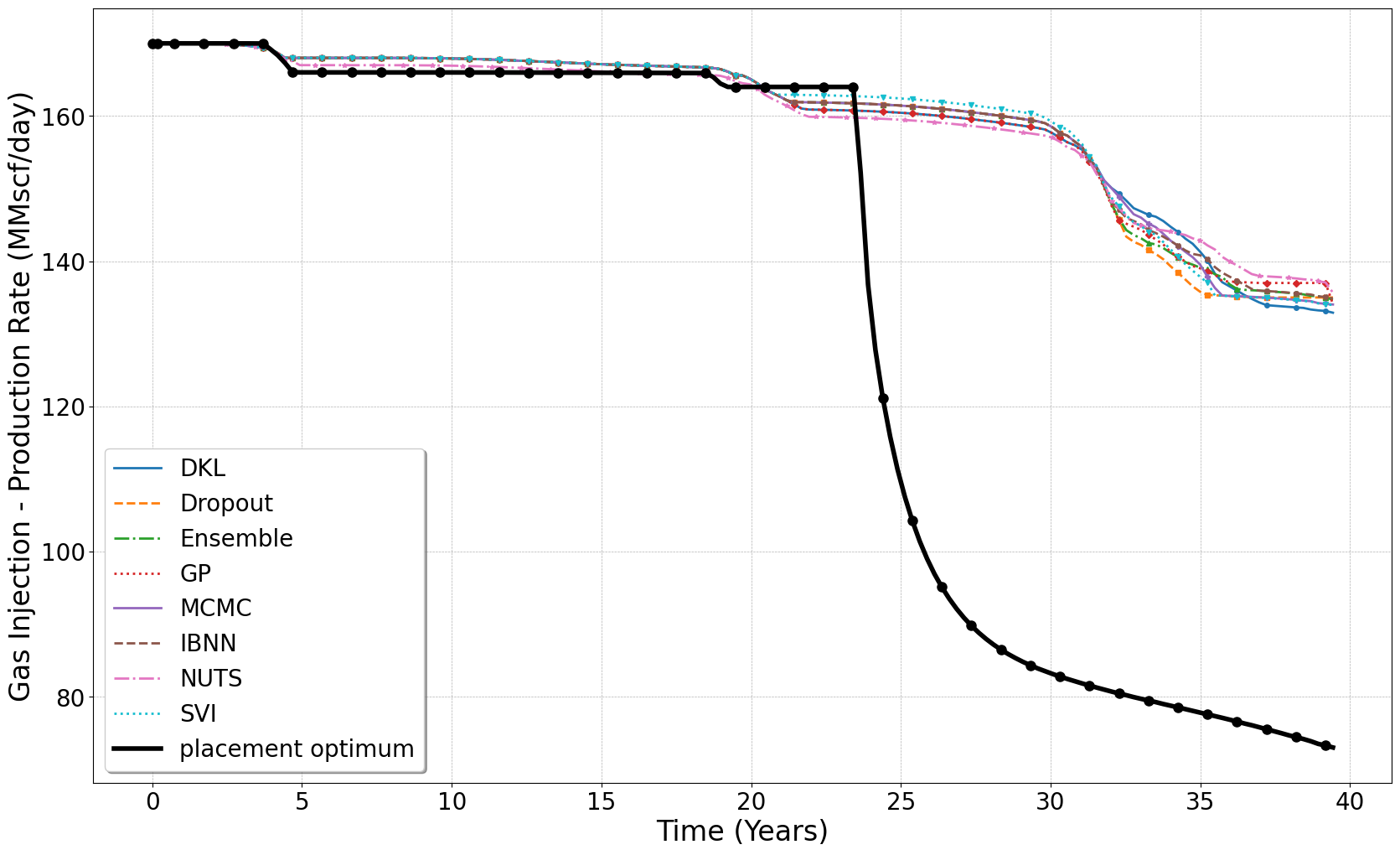}
\caption{Sequestration rate \label{fig7}}
\end{figure}

\begin{figure}[H]
\includegraphics[width=0.7\linewidth,keepaspectratio]{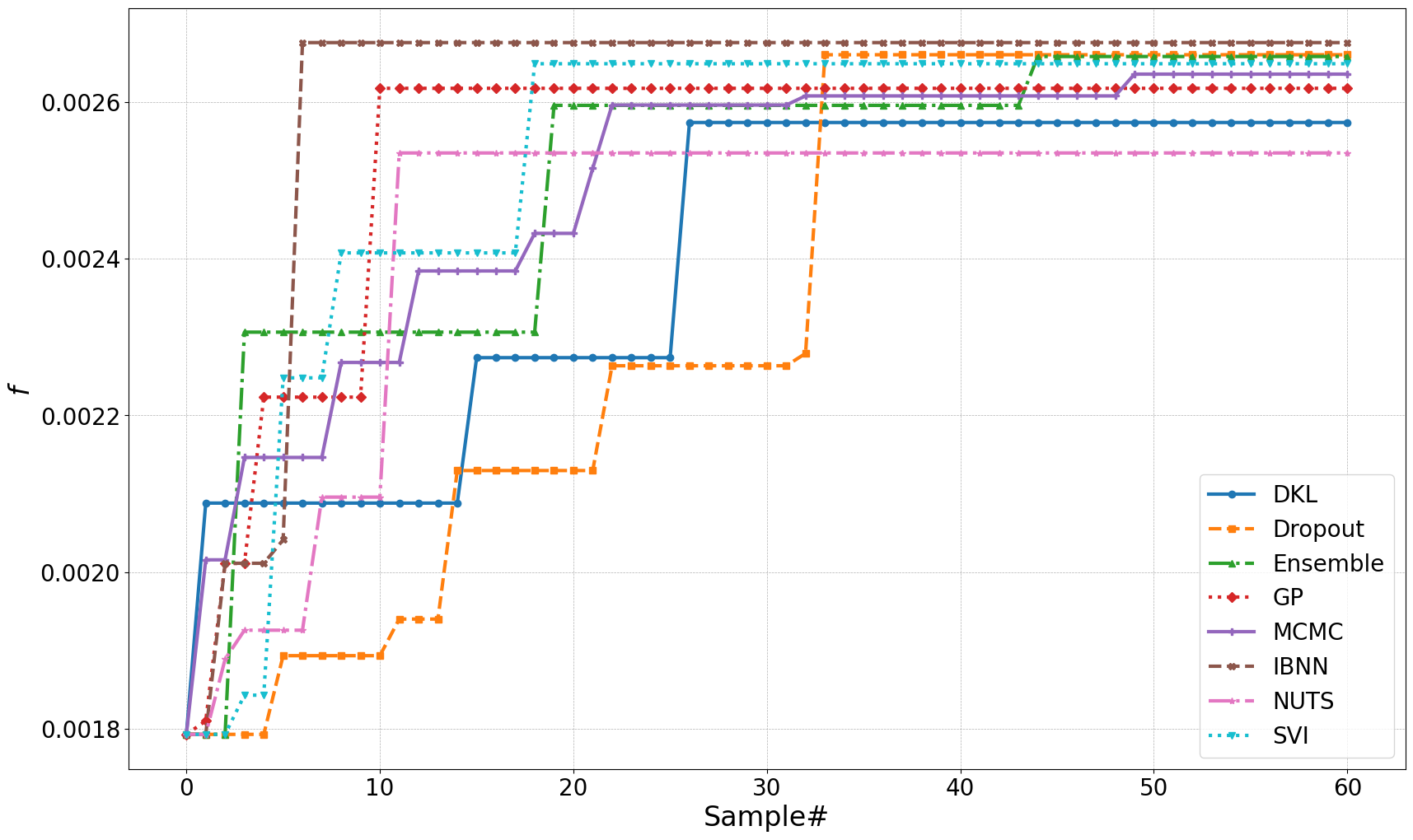}
\caption{Maximum objective function value vs iteration number\label{fig8}}
\end{figure}

We begin with the first variation of Case study 1. The optimum sequestration rate from the application of BO to each model is shown in Figure \ref{fig7} along with the placement optimum case to facilitate the comparison. As already mentioned, the optimization has been able to maintain the injection rate at the target value throughout the lifespan of the project, therefore the sequestration rate is shown instead, i.e. injected minus produced CO$_2$, for the comparison to be meaningful. Since breakthrough inevitably occurs, the expectation is for the models to be able to keep this sequestration rate constant for as long as possible and as close as possible to the target value. The goal is to delay the onset of the sequestration-rate decline and reduce its steepness relative to the placement optimum case.
The results in Figure \ref{fig7} indicate that, compared to the benchmark placement optimum case, the solutions proposed by the optimizer do indeed fix the aforementioned issues. In the benchmark case, the sequestration rate drop occurs right after 23 operational years. In comparison, most optimized solutions don't experience this drop up until a decade later, at 31-32 years. All models were able to find the optimal solutions that manage to keep the sequestration rate above 75\% of the initial target even as late as the end of the operation, in contrast to the placement optimum, where the sequestration rate dropped to below 50\% by the end of the operation. Notably, all models were able to eventually sequester around 2.3 Tscf of CO$_2$ out of 2.45 theoretically possible as opposed to the placement optimum case where 1.9 Tscf were sequestered. In Figure \ref{fig8} the maximum objective function value derived per model is plotted to demonstrate the evolution across sampled values in subsequent BO iterations. 
For example, IBNN found improvements over the first two BO iterations but was unable to further improve its predictions. This suggests that the value found lies near a (global or local) optimum one rather than an inability of BO in general. On the other hand, Dropout was consistently finding points with better evaluations and eventually converged close to the optimum but, it was the slowest to reach its best evaluation and finished in second place. The third highest value was found by the Ensemble model. Like IBNN, GP also didn't achieve further improvements after the third iteration and its max value is lower than the highest one observed. This is notable given that GP-based BO is often strong in low-dimensional settings. The MCMC model also managed to improve throughout BO iterations and eventually converged close to the maximum. The DKL model performed slightly worse compared to the others. As for the Pyro models, NUTS achieved most of its improvement at the first iterations but was unable to improve after, ranking it last in terms of max value produced. On the other hand, SVI consistently improved fast and was able to approach the overall maximum value more closely.

\subsubsection{Variation 2}
\begin{figure}[H]
\includegraphics[width=0.7\linewidth,keepaspectratio]{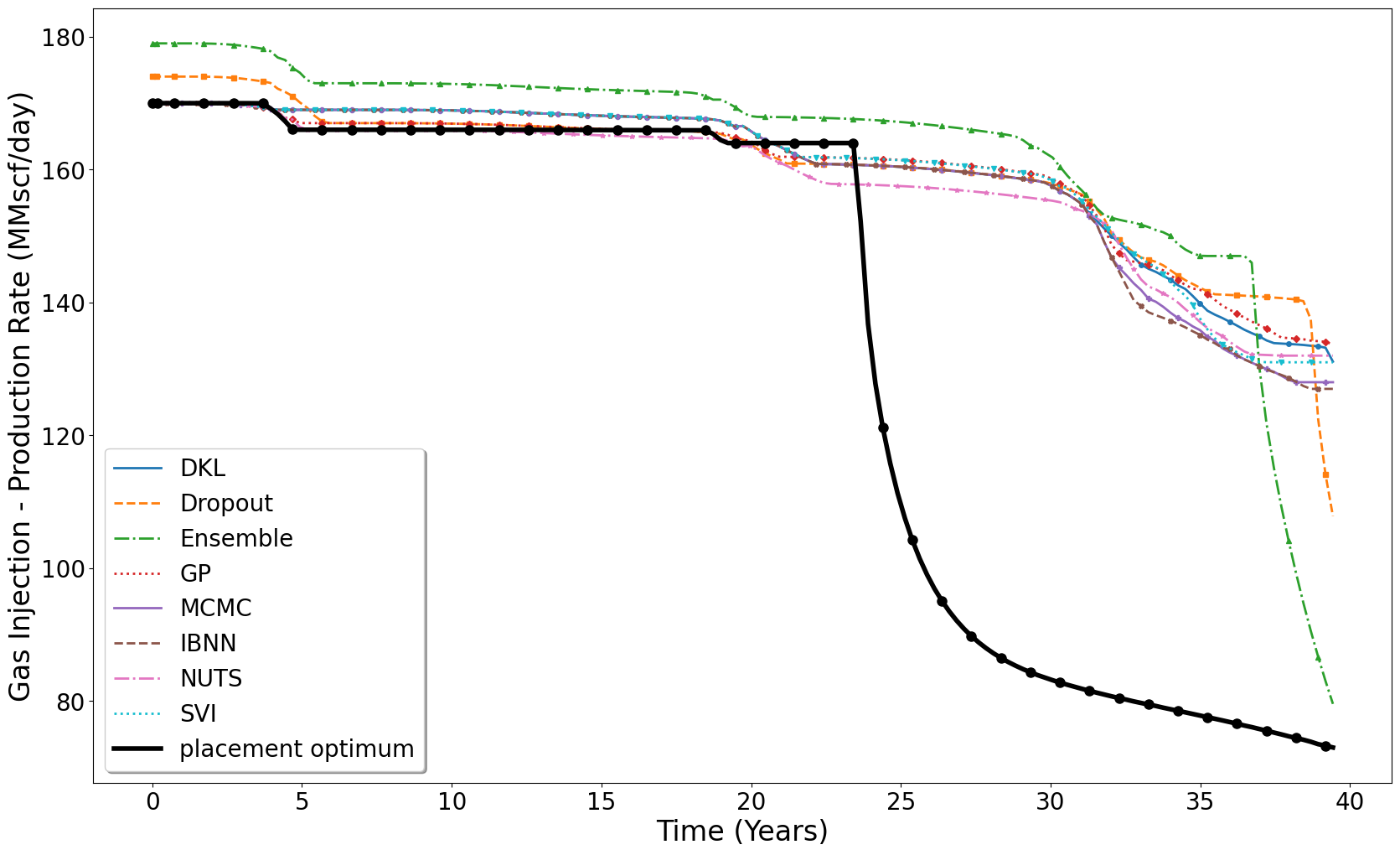}
\caption{Sequestration rate \label{fig9}}
\end{figure}

In the second variation, results are more varied as depicted in Figure \ref{fig9} which is the plot of the sequestration rate of the best solutions of all models selected by maximizing the second objective. In this variation models were free to explore higher injection target rates, but most models did not identify improved hypervolume by increasing the injection target, despite the added total-sequestration objective. 
The placement optimum case is still the first one to have its sequestration rate drop off rapidly. All models besides Ensemble and Dropout converged to a solution that produced a similar sequestration profile as variation one. Ensemble and Dropout however, explored towards higher injection rate solutions and in both cases the steep late-time decline reduces the dominated hypervolume, making these solutions suboptimal under the chosen reference point.
\begin{figure}[H]
\includegraphics[width=0.7\linewidth,keepaspectratio]{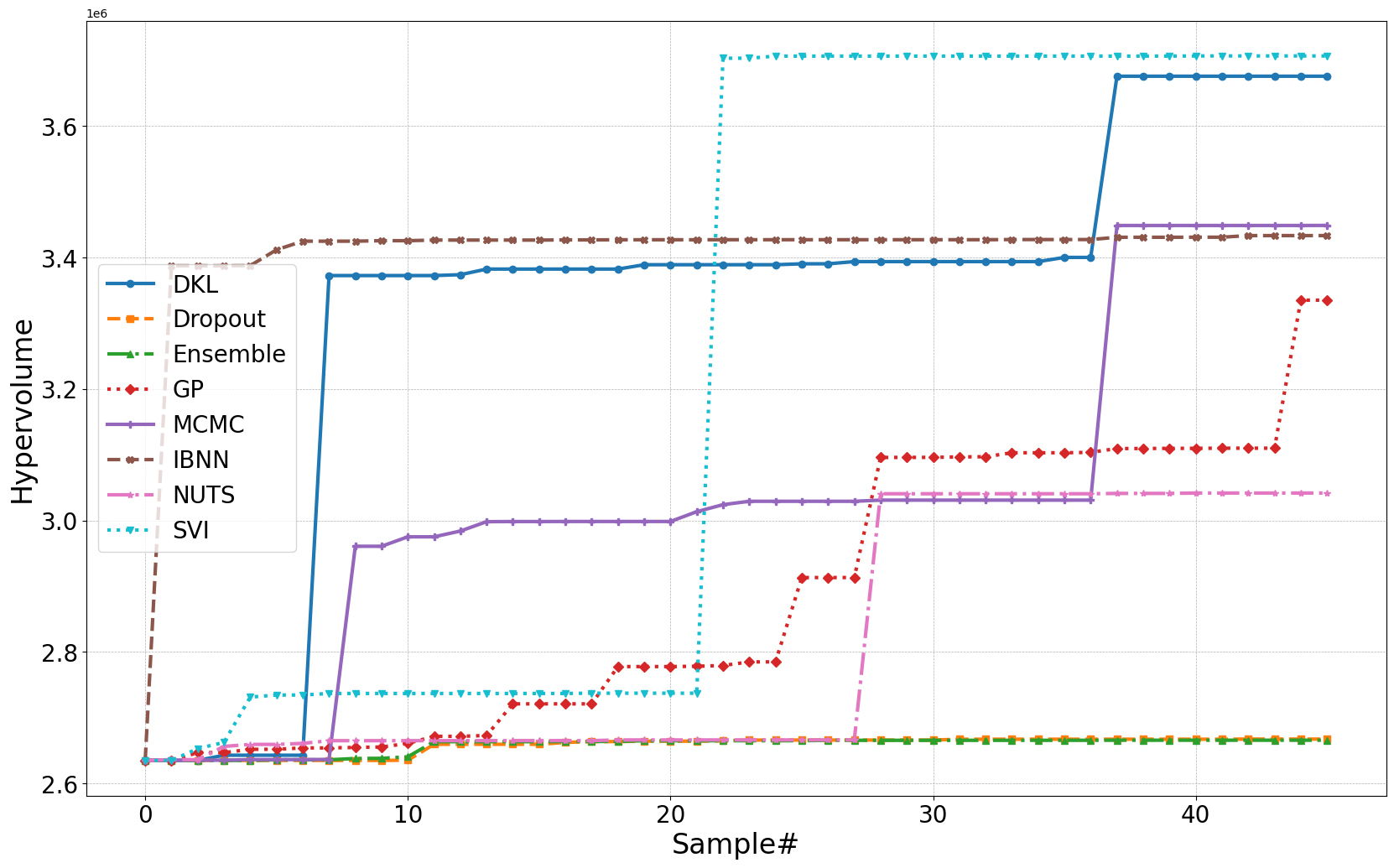}
\caption{Hypervolume vs iteration number\label{fig10}}
\end{figure}
Figure \ref{fig10} presents the evolution of the hypervolume for all surrogate models. The Ensemble and Dropout models have slightly improved only at early BO iterations but due to their exploring different optimal solutions, they finished last in terms of hypervolume maximization. In contrast and similarly to the first variation, IBNN improved significantly at an early BO iteration and then struggled for further improvements. NUTS improved slightly at first and significantly only once at a later iteration. MCMC and GP improved steadily throughout the BO iterations. SVI achieved the best result overall, closely followed by DKL.

\subsection{Case study 2}
In this case study, the optimization problem involves a high number of variables and four competing objective functions. Notably, the first and third objectives are negatively correlated with the second and fourth. Specifically, the first objective evaluates the safety of CO$_2$ sequestration, while the third measures the decline in injection rate, primarily due to overpressurization. These objectives tend to be maximized when the injection rate is low. In contrast, the second and fourth objectives assess total sequestration capacity and economic profitability, both of which benefit from higher injection rates. Given the increased complexity, we increased the computational budget relative to Case study 1 and repeated the optimization across eight independent trials. Each trial consisted of 120 BO iterations with a batch of 5 candidate evaluations per iteration.

\begin{figure}[H]
\includegraphics[width=0.7\linewidth,keepaspectratio]{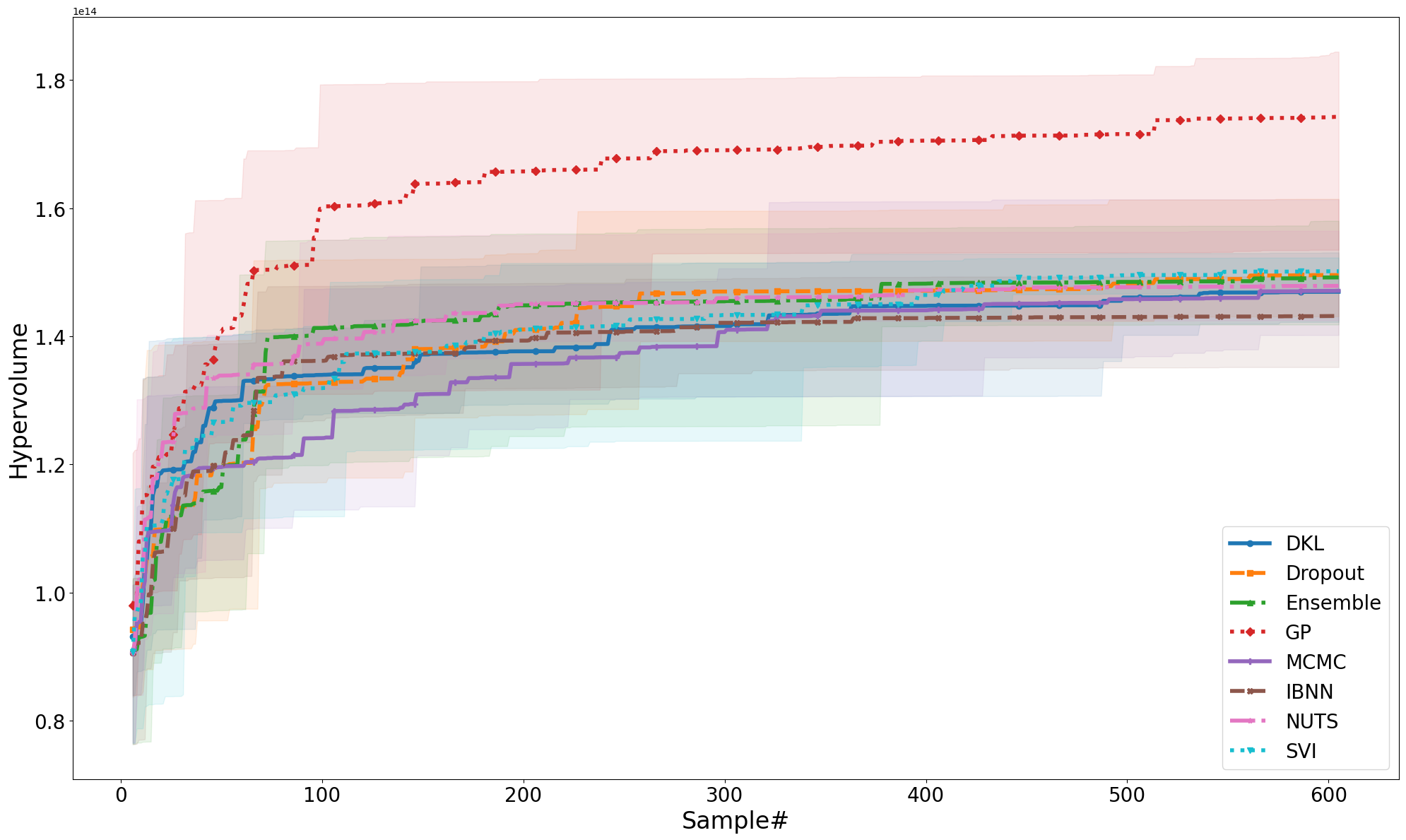}
\caption{Hypervolume vs iteration number\label{fig15}}
\end{figure}

Figure \ref{fig15} presents the evolution of the hypervolume measure at each iteration, representing the hypervolume of the current set of all non-dominated solutions found up to that point. The characteristic behavior of the hypervolume curves is evident: an initially low value, followed by a rapid increase during the early iterations and eventually a plateau where improvements become marginal. Among the models tested, GP achieved the highest average hypervolum in this benchmark. In contrast, all other models plateau at significantly lower values. Regarding convergence speed, GP, Ensemble and NUTS converge more rapidly, whereas Dropout, MCMC and DKL exhibit slower convergence on average.

\begin{figure}[H]
\includegraphics[width=\linewidth,keepaspectratio]{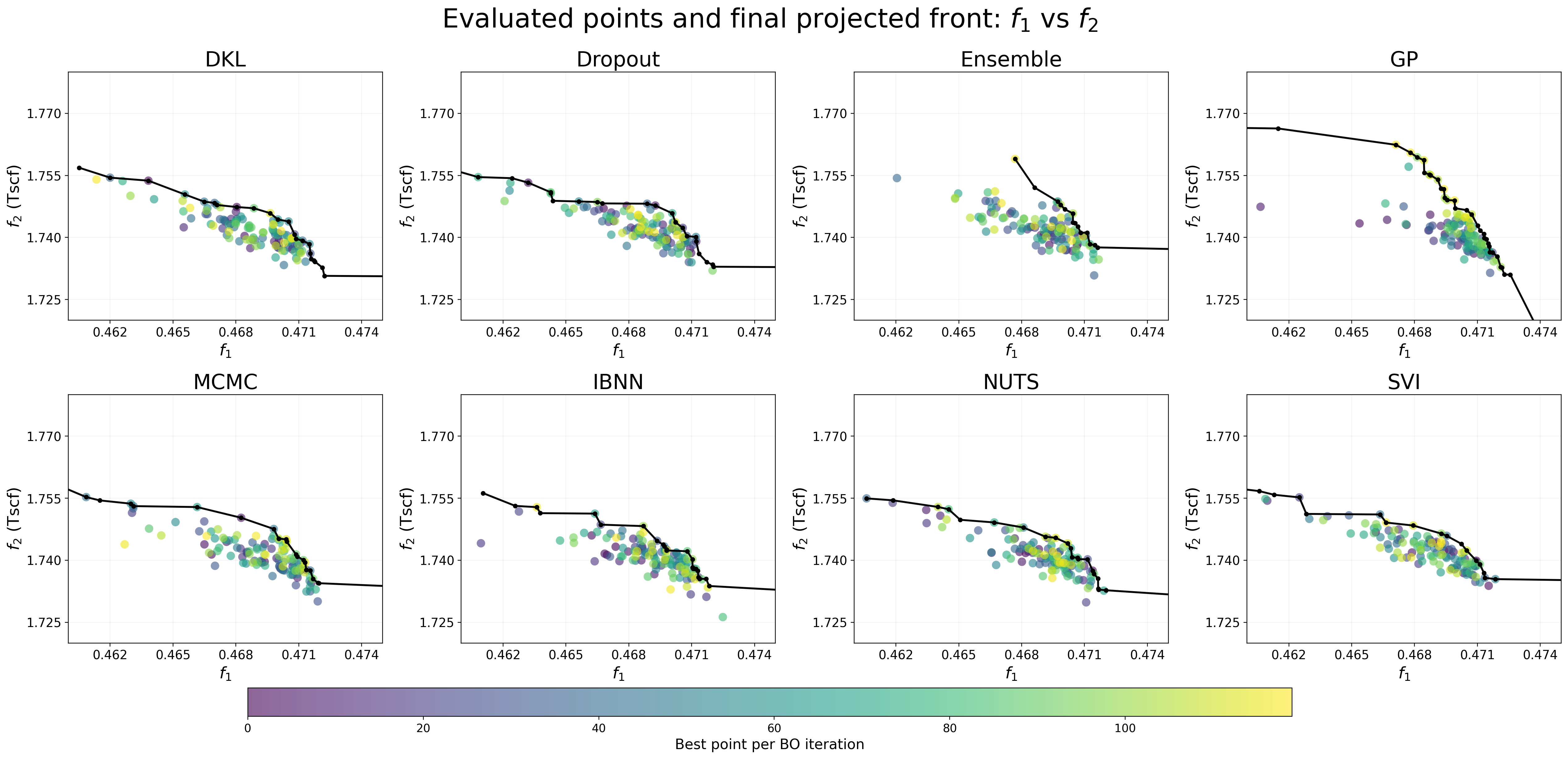}
\caption{Evaluated solutions and final projected fronts for Case Study 2 in the $f_1$--$f_2$ plane.\label{fig:case2_f1f2}}
\end{figure}

\begin{figure}[H]
\includegraphics[width=\linewidth,keepaspectratio]{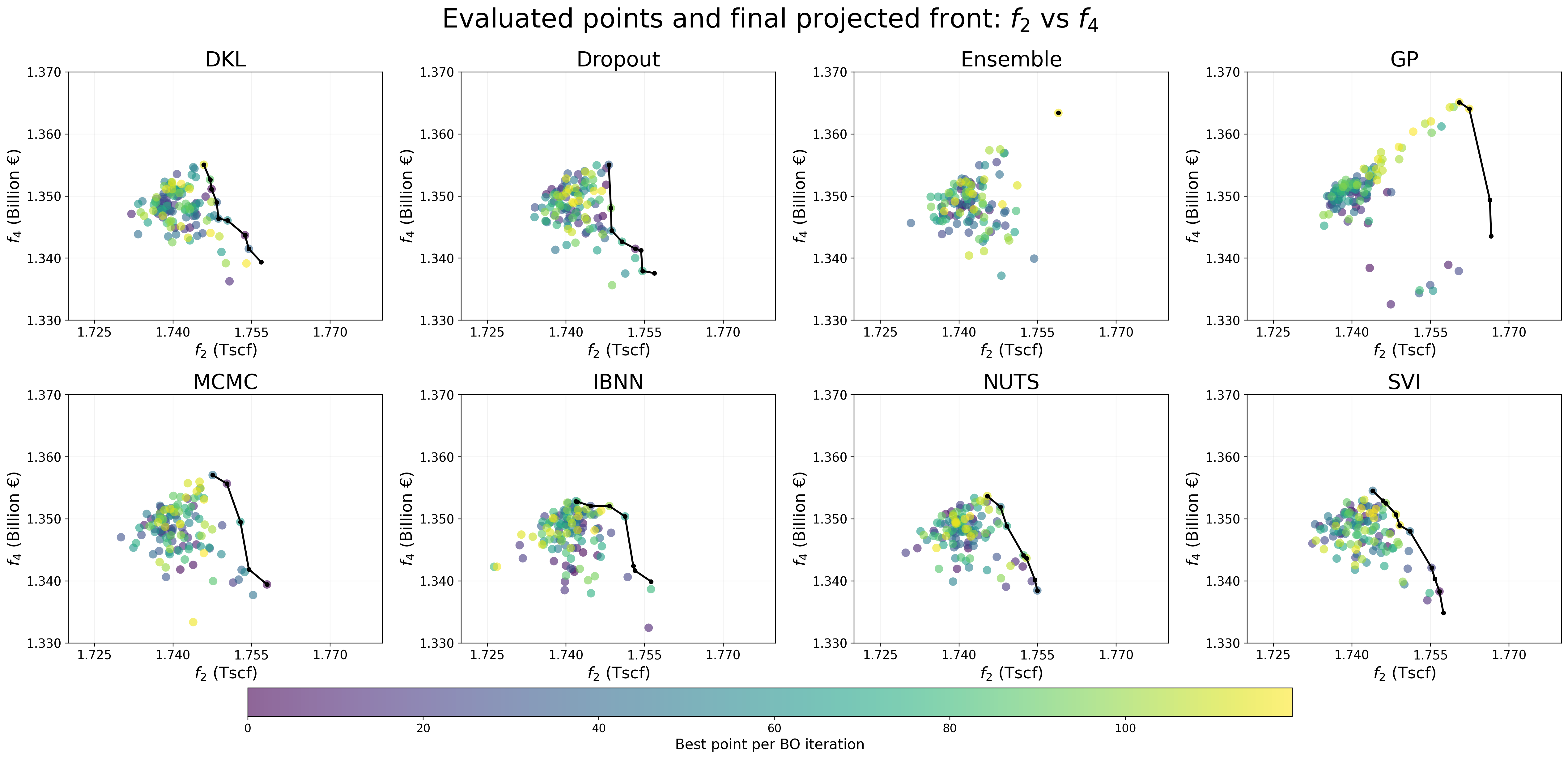}
\caption{Evaluated solutions and final projected fronts for Case Study 2 in the $f_2$--$f_4$ plane.\label{fig:case2_f2f4}}
\end{figure}

To complement the hypervolume-based comparison, Figures \ref{fig:case2_f1f2} and \ref{fig:case2_f2f4} show projected views of the evaluated solutions and the final projected fronts for two representative objective pairs. These plots are not meant to replace the hypervolume metric, but rather to provide an intuitive view of the trade-off structure captured by the optimization process.

Figure \ref{fig:case2_f1f2} highlights the expected tension between trapping-related performance and total sequestration capacity. The projected fronts illustrate that increasing $f_2$ generally comes at the expense of lower values of $f_1$, reflecting the underlying trade-off between aggressive injection-driven storage and more conservative, safety-oriented operating conditions. The evaluated points also show that most models progressively concentrate their search near the final projected front, indicating that BO is able to identify and refine the relevant trade-off region over time.

In contrast, Figure \ref{fig:case2_f2f4} shows that total sequestration and NPV are much more closely aligned. The projected fronts lie along a relatively narrow band, indicating that improvements in stored CO$_2$ volume are generally accompanied by improved economic performance under the present simplified NPV formulation. This helps explain why the second and fourth objectives exhibit similar behavior in several of the supplementary diagnostics, and why the main multi-objective distinction in this case is driven more strongly by the trade-off between sequestration-oriented and safety-oriented objectives than by a direct conflict between sequestration and profitability.

Objective-wise best-value tables and convergence plots for Case Study 2 are retained in Appendix A only as supplementary diagnostic material.

\section{Discussion}
\label{sec5}
In this work, we benchmark BNN-family stochastic surrogates against GP baselines for Bayesian optimization of CO$_2$ storage operations. At a high level, the optimization results are operationally meaningful and improve upon the benchmark schedules. In Case study 1, by simply optimizing the rates without adding any sort of extra operational cost, the operator is now able to steadily inject the target amount of CO$_2$ resulting in much higher total sequestration than the benchmark case. In Case study 2, under a reduced producer configuration (fewer wells), which reflects CAPEX constraints, the optimizer was able to sequester up to 91\% of the CO$_2$ mass achieved in the benchmark case, although the total CO$_2$ mass sequestered accounted for only one of the four objective functions. Similarly, the optimal NPV computed for most models was as high as 92\% of the benchmark, despite the fact that five less wells were utilized in the injection phase.
\\\\
These outcome-level improvements provide context, but the primary aim is to compare surrogate behavior within a consistent BO protocol. The most important and most difficult is to compare the ability of the various stochastic models to act as suitable surrogate models for BO. GPs are often used as the default approach due to ease of implementation and ability to be customized by a kernel function to shape the posterior probability distribution as needed. BNNs constitute an attractive alternative because they can learn flexible, potentially non-stationary representations that may be beneficial when standard stationary-kernel GP assumptions are restrictive. Within the tested budgets and hyperparameter settings, no surrogate family dominates across all cases. On a positive note, their performance was comparable to GPs in most cases and outperformed them at times. In Case study 1, GPs were expected to outperform BNNs due to the small number of decision variables. GPs proved able to only find more improvements in the objective function than BNNs in the second variation of Case study 1 but not in the first. Furthermore, they were never able to clearly outperform BNNs in terms of finding a better overall optimum. On the other hand, in Case study 2, BNNs were expected to take the upper hand due to the large increase in decision variables and the use of multiple objectives with completely different scales. In this benchmark, the GP baseline achieved the strongest average performance in Case study 2.
\\\\
On another note, it is important to question whether or not utilizing approximate rather than exact Bayesian inference is preferable when incorporated to neural networks. The NUTS and SVI models training took much more time than other methods due to a more intricate sampling process. However, their performance proved to be similar to that of approximate Bayesian methods such as Dropout, Ensemble and DKL. The higher inference cost did not translate into consistently stronger BO outcomes under the fixed evaluation budgets used here. Nevertheless, this analysis is not fully comprehensive as the models are very sensitive to the choice of the hyperparameters. A more exhaustive hyperparameter and architecture sensitivity study would help clarify when expensive inference yields measurable gains. Another limitation of the present benchmark is that acquisition functions were not varied systematically; instead, a fixed qEI/qEHVI framework was retained in order to compare surrogate families under a controlled BO protocol.
\\\\
Another important reason for the models not exhibiting pronounced differences in their performance is the high non-linearity between the decision variables and the objective function output. The decision variables utilized in the optimization process are the rates that each well or group is targeted to produce/inject, not necessarily the ones that will actually be implemented. For example, when a well is set to operate at a production rate such that the bottom hole pressure limit is reached early, the control policy will force the actual production to significantly drop giving a non-linear response to the objective function. This shouldn't be a problem for the surrogates in this work as they are capable of coping with non-linearity. Non-linearity however is enhanced not only by bottomhole pressure constraints but also to acceptable CO$_2$ recycling rate constraint as well as the combined effects that other impose on its performance. The non-linearity may be further enhanced to the point where correlations are weak or inconsistent between the decision variables and the observed output. This is much more strongly demonstrated in Case study 2, where the decision variables at the later time steps are very sensitive to the selection of the decision variables at earlier time steps when it comes to their effect on the system response. Nevertheless, it should be noted that, apart from the performance of the optimization methods, this behavior is a direct result of this controlled synthetic benchmark in comparison to areally extended ones typically considered in CCS operations. A further limitation of the present study is that all surrogate models treat the reservoir simulator as a black box and do not explicitly incorporate the underlying spatial-temporal physics, which may reduce data efficiency in highly nonlinear CCS systems. At present, multi-fidelity Bayesian optimization appears to be a particularly promising direction for further investigation \cite{do2025multifidelity}. For example, in the subsurface engineering domain, physics-informed machine-learning frameworks have recently been proposed for reservoir-connectivity identification and production forecasting in CO$_2$-EOR systems, while hierarchical spatio-temporal surrogate architectures and attention-based spatial-temporal models have been used to better capture long-range flow dynamics in CO$_2$ sequestration and CO$_2$-WAG settings \cite{nagao2024physics,lin2024towards,xu2024advanced,na2024application,tang2025graph}. In parallel, smart proxy and graph-based surrogate approaches have shown potential for accelerating CCS-related optimization tasks. These developments indicate that future work could extend the present benchmark toward physics-informed, graph-based, and multi-fidelity BO surrogates that encode more of the governing flow structure.

\section{Conclusions}
\label{sec6}
BO offers a data-efficient framework for CCS operational control problems where each evaluation requires a full-physics simulation. Following the work of Li \cite{li2023study} it has been much easier recently to apply BO and even evaluate the performance of various robust BNN surrogates. Research in BO has started to take off recently and this work contributes a controlled benchmark in reservoir-engineering CCS settings for comparing stochastic surrogate families within BO. The method proved to be a very powerful process in the single as well as the multi-objective and many decision variables cases, that CCS projects usually constitute. It was shown to provide realistic solutions and greatly improve on the benchmark reduced placement optimum case. A key outcome is the improvement in NPV relative to the reduced-placement benchmark under optimized controls. The NPV can be the metric that makes or breaks a CCS project, as it can give the initial incentive to operators and stakeholders to take up a project and allocate resources towards it. In this work the simplified NPV formulation is a limitation as it depends on assumed CO$_2$ storage pricing, which is subject to substantial regulatory and market uncertainty; consequently, the corresponding results should be interpreted as indicative rather than as precise economic forecasts.
\\\\
This study provides evidence on the surrogate models' ability to find optimal values despite the limitations of the framework discussed in Section \ref{sec5}. Future work should prioritize (i) better-correlated decision parameterizations and constraint formulations, and (ii) scaling to larger, more heterogeneous models; additional extensions include detecting decision variables that better correlate to output performance, utilizing larger models for better correlation of system response and decision variables, combining well placement and well rates as decision variables simultaneously, adding models of larger complexity and more detailed geomechanical constraints and last but not least, utilizing optimization algorithms in a closed loop setting possibly by Reinforcement Learning \cite{sutton2018reinforcement} algorithms that provide very suitable frameworks for this task.

 \section*{Credit author statement}
 Conceptualization, S.F. and V.G.; methodology, S.F.; software, S.F.; validation, S.F. and V.G.; resources, S.F.; writing---original draft preparation, S.F.; writing---review and editing, S.F., V.G.; visualization, S.F. All authors have read and agreed to the final version of the manuscript.

\section*{Declaration of competing interest} The authors declare that they have no known competing financial interests or personal relationships that could have appeared to influence the work reported in this paper.

\section*{Funding} The resources were granted with the support of GRNET as part of the project StorageCO$_2$.
The publication of the article in OA mode was financially supported by HEAL-Link

\section*{Code Availability}
Code used in this study, including the Bayesian optimization framework, surrogate-model implementations, experiment configurations, and selected archived outputs, is publicly available at \url{https://github.com/flammmes/bnn-bo}. Reservoir simulation deck files (\texttt{.DATA} and related simulator files) are not distributed due to confidentiality restrictions.

\newpage

\appendix
\section{Supplementary objective-wise diagnostics for Case Study 2}

For completeness, this appendix reports supplementary objective-wise diagnostics for Case Study 2, including benchmark objective values, objective-wise best-value tables, and convergence plots for each objective.

The results for the four objectives in both the reduced placement optimum and placement optimum benchmark cases are presented in Table \ref{tab2}. As a reminder, both cases were run with a fixed target injection rate of 170 MMscf/day.

\begin{table}[H]
\centering
\caption{Objective function results on benchmark cases \label{tab2}}
\begin{tabular}{l|cccc}
\toprule
 & $f_1$ & $f_2\;(Tscf)$ & $f_3$ & $f_4$ (Billion \euro)\\
\midrule
placement optimum & 0.3885 & 1.9292 & 0.0012 & 1.4673 \\
reduced placement optimum & 0.4831 & 1.4062 & 0.0006 & 1.1511 \\
\bottomrule
\end{tabular}
\end{table}

When the number of producers is reduced, the last three objective function values decline significantly. For instance, the NPV of the project decreases by 300 million \euro, a loss that far exceeds the cost savings from drilling fewer producers. Additionally, total CO$_2$ sequestration drops substantially, shortening the overall project lifespan---a trend already illustrated in Figure \ref{fig6}. The only exception is the first objective function, which increases as expected. With an early decline in injection rate and a corresponding rise in pressure, the CO$_2$ plume becomes immobile more quickly, leading to a higher proportion of dissolution relative to newly injected CO$_2$.

Since this is a multi-objective optimization problem with no single globally optimal solution, the tables below are provided only as supplementary objective-wise diagnostics. Table \ref{tab7} reports the solutions from each model that maximized the first objective. Following the same approach, Tables \ref{tab8}, \ref{tab9} and \ref{tab10} present the best solutions corresponding to the second, third and fourth objectives, respectively.

\begin{table}[H]
\centering
\caption{Supplementary objective-wise results for Case Study 2 according to the first objective function \label{tab7}}
\begin{tabular}{l|cccc}
\toprule
 & $f_1$ & $f_2\;(Tscf)$ & $f_3$ & $f_4$ (Billion \euro)\\
\midrule
DKL & 0.5783 ± 0.0090 & 0.6022 ± 0.0405 & 0.6269 ± 0.5149 & 0.5560 ± 0.0323 \\
Dropout & 0.5840 ± 0.0136 & 0.5816 ± 0.0567 & 0.5016 ± 0.5328 & 0.5400 ± 0.0448 \\
Ensemble & 0.5755 ± 0.0134 & 0.6245 ± 0.0565 & 0.3769 ± 0.5160 & 0.5750 ± 0.0433 \\
GP & 0.5266 ± 0.0466 & 1.1171 ± 0.5277 & 0.3754 ± 0.5172 & 0.9220 ± 0.3665 \\
MCMC & 0.5799 ± 0.0191 & 0.6038 ± 0.0797 & 0.4220 ± 0.4941 & 0.5580 ± 0.0624 \\
IBNN & 0.5888 ± 0.0171 & 0.5632 ± 0.0743 & 0.2518 ± 0.4618 & 0.5255 ± 0.0609 \\
NUTS & 0.5839 ± 0.0141 & 0.5850 ± 0.0611 & 0.2511 ± 0.4622 & 0.5436 ± 0.0497 \\
SVI & 0.5924 ± 0.0131 & 0.5504 ± 0.0563 & 0.2509 ± 0.4624 & 0.5158 ± 0.0468 \\
\bottomrule
\end{tabular}
\end{table}

Compared to the benchmark cases, all models achieved significantly higher values for the first objective (Table \ref{tab7}), while the values for the other three objectives were considerably lower. This suggests that these solutions correspond to scenarios where the injection rate was initially average, but brine production remained minimal. As a result, pressure builds up rapidly, causing the injection rate to decline quickly. This leads to the CO$_2$ plume becoming immobile, which is interpreted as trapped by the OPM simulator. An exception to this trend is observed in the Ensemble and NUTS models, where the third objective reaches its maximum possible value of 1. However, if the injection rate were maintained constant over the entire 40-year period, much higher values for both NPV and total sequestration would be expected. A plausible explanation is that in these cases, the injection rate remains low, the production rate is high, and the recycling constraint is significantly relaxed.

\begin{table}[H]
\centering
\caption{Supplementary objective-wise results for Case Study 2 according to the second objective function \label{tab8}}
\begin{tabular}{l|cccc}
\toprule
 & $f_1$ & $f_2\;(Tscf)$ & $f_3$ & $f_4$ (Billion \euro)\\
\midrule
DKL & 0.4605 ± 0.0012 & 1.7576 ± 0.0017 & 0.0011 ± 0.0002 & 1.3396 ± 0.0020 \\
Dropout & 0.4601 ± 0.0007 & 1.7568 ± 0.0013 & 0.0013 ± 0.0002 & 1.3387 ± 0.0012 \\
Ensemble & 0.4611 ± 0.0027 & 1.7570 ± 0.0014 & 0.0011 ± 0.0003 & 1.3415 ± 0.0090 \\
GP & 0.4590 ± 0.0006 & 1.7605 ± 0.0042 & 0.0016 ± 0.0001 & 1.3392 ± 0.0033 \\
MCMC & 0.4598 ± 0.0008 & 1.7574 ± 0.0012 & 0.0013 ± 0.0002 & 1.3379 ± 0.0019 \\
IBNN & 0.4598 ± 0.0011 & 1.7576 ± 0.0017 & 0.0013 ± 0.0002 & 1.3377 ± 0.0029 \\
NUTS & 0.4611 ± 0.0009 & 1.7548 ± 0.0008 & 0.0012 ± 0.0002 & 1.3385 ± 0.0024 \\
SVI & 0.4603 ± 0.0008 & 1.7569 ± 0.0015 & 0.0013 ± 0.0002 & 1.3391 ± 0.0027 \\
\bottomrule
\end{tabular}
\end{table}

For the optimal values obtained with respect to the second objective (Table \ref{tab8}), it is evident that the solutions are realistic and show significant improvement over the reduced placement optimum case. Although none of the models achieve the maximum sequestration of 1.92 Tscf observed in the placement optimum case, all models reach approximately 1.76 Tscf, which is a substantial increase from the initial 1.4 Tscf.

\begin{table}[H]
\centering
\caption{Supplementary objective-wise results for Case Study 2 according to the third objective function \label{tab9}}
\begin{tabular}{l|cccc}
\toprule
 & $f_1$ & $f_2\;(Tscf)$ & $f_3$ & $f_4$ (Billion \euro)\\
\midrule
DKL & 0.5532 ± 0.0250 & 0.7345 ± 0.1457 & 1.0000 ± 0.0000 & 0.6575 ± 0.1084 \\
Dropout & 0.5519 ± 0.0306 & 0.7454 ± 0.1749 & 1.0000 ± 0.0000 & 0.6645 ± 0.1318 \\
Ensemble & 0.5544 ± 0.0218 & 0.7235 ± 0.1233 & 0.8764 ± 0.3497 & 0.6491 ± 0.0923 \\
GP & 0.5047 ± 0.0447 & 1.2060 ± 0.4825 & 0.6257 ± 0.5166 & 0.9781 ± 0.3210 \\
MCMC & 0.5492 ± 0.0214 & 0.7511 ± 0.1283 & 1.0000 ± 0.0000 & 0.6703 ± 0.0950 \\
IBNN & 0.5434 ± 0.0149 & 0.7746 ± 0.0762 & 0.8763 ± 0.3498 & 0.6882 ± 0.0551 \\
NUTS & 0.5595 ± 0.0228 & 0.6932 ± 0.1020 & 1.0000 ± 0.0000 & 0.6262 ± 0.0809 \\
SVI & 0.5556 ± 0.0083 & 0.7122 ± 0.0419 & 0.7515 ± 0.4601 & 0.6425 ± 0.0331 \\
\bottomrule
\end{tabular}
\end{table}

For the results corresponding to the third objective, all models successfully reached the maximum possible value (Table \ref{tab9}). The underlying explanation is similar to what was observed for two models in Table \ref{tab7}: the injection rate never declines due to a high recycling rate. These cases are characterized by low injection rates and high production rates, with a very lenient gas production constraint. Consequently, despite continuous injection, both NPV and total sequestration remain low.

\begin{table}[H]
\centering
\caption{Supplementary objective-wise results for Case Study 2 according to the fourth objective function \label{tab10}}
\begin{tabular}{l|cccc}
\toprule
 & $f_1$ & $f_2\;(Tscf)$ & $f_3$ & $f_4$ (Billion \euro)\\
\midrule
DKL & 0.4699 ± 0.0004 & 1.7467 ± 0.0021 & 0.0004 ± 0.0000 & 1.3565 ± 0.0022 \\
Dropout & 0.4697 ± 0.0008 & 1.7470 ± 0.0025 & 0.0004 ± 0.0000 & 1.3561 ± 0.0009 \\
Ensemble & 0.4696 ± 0.0011 & 1.7491 ± 0.0065 & 0.0004 ± 0.0000 & 1.3580 ± 0.0036 \\
GP & 0.4696 ± 0.0010 & 1.7501 ± 0.0069 & 0.0004 ± 0.0000 & 1.3592 ± 0.0043 \\
MCMC & 0.4700 ± 0.0004 & 1.7468 ± 0.0021 & 0.0004 ± 0.0000 & 1.3567 ± 0.0011 \\
IBNN & 0.4700 ± 0.0006 & 1.7447 ± 0.0022 & 0.0004 ± 0.0000 & 1.3543 ± 0.0010 \\
NUTS & 0.4698 ± 0.0006 & 1.7452 ± 0.0024 & 0.0004 ± 0.0000 & 1.3546 ± 0.0007 \\
SVI & 0.4699 ± 0.0003 & 1.7446 ± 0.0008 & 0.0004 ± 0.0000 & 1.3549 ± 0.0006 \\
\bottomrule
\end{tabular}
\end{table}

Similarly to the second objective, the best solutions for the fourth objective (Table \ref{tab10}) exhibit a more conventional behavior compared to those of the first and third objectives. The NPV values increase by 200 million \euro relative to the benchmark reduced placement optimum case. Although the NPV remains lower than that of the placement optimum case, the trade-off between achieving a lower NPV with only three wells drilled versus a higher NPV with eight wells drilled can be justified more convincingly.

Notably, by comparing the best solutions obtained for the second and fourth objectives, a key distinction emerges: in the latter case, the injection rate declines more rapidly due to reduced brine production, as it incurs additional costs. In contrast, for the second objective, the injection rate typically decreases more gradually since brine withdrawal is not a limiting factor. This distinction highlights that while both cases yield similar solutions in terms of overall performance, their solution profiles differ significantly, aligning with the priorities of their respective optimization objectives.

\begin{figure}[H]
\includegraphics[width=0.7\linewidth,keepaspectratio]{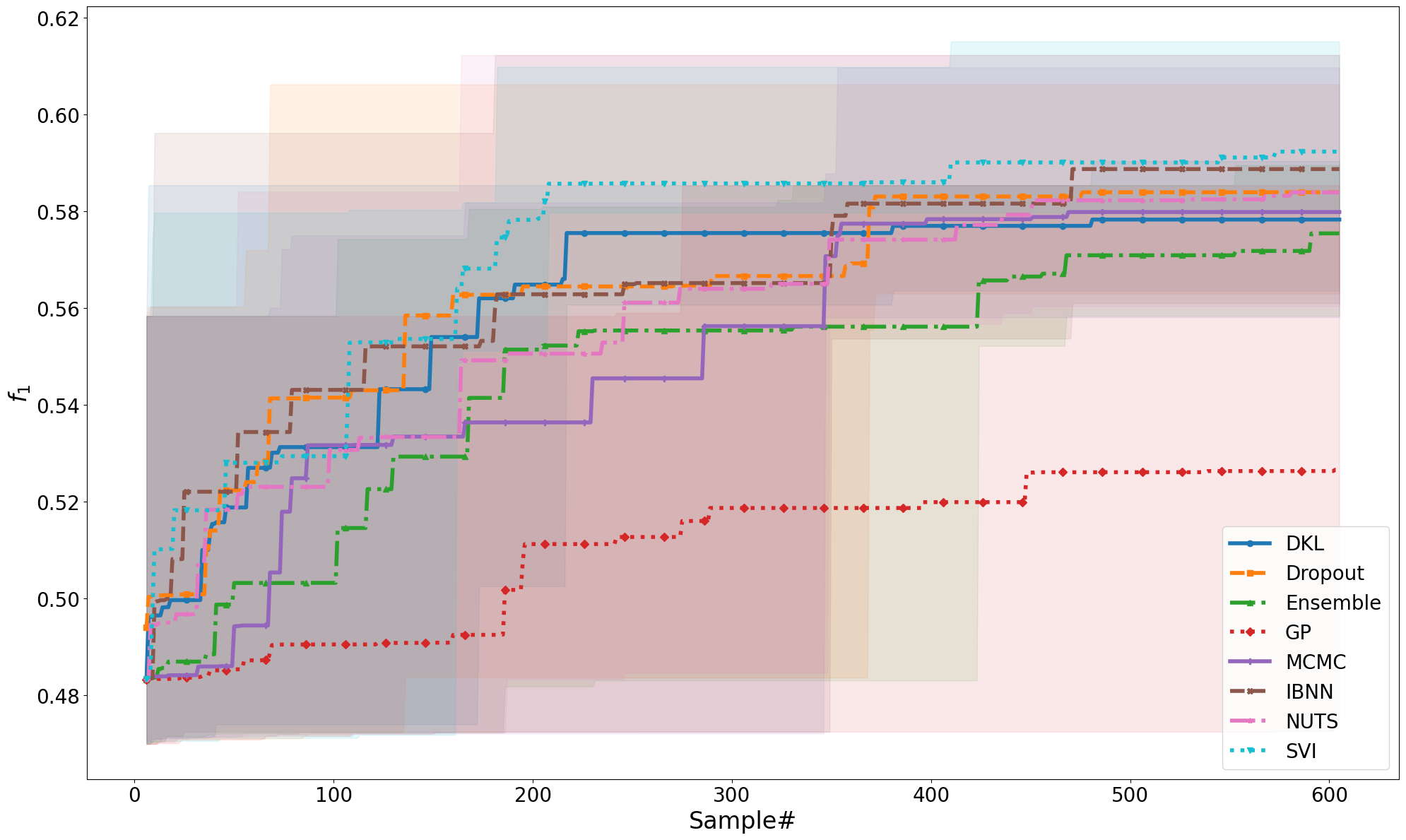}
\caption{Supplementary convergence of the maximum first objective value vs iteration number\label{fig11}}
\end{figure}

\begin{figure}[H]
\includegraphics[width=0.7\linewidth,keepaspectratio]{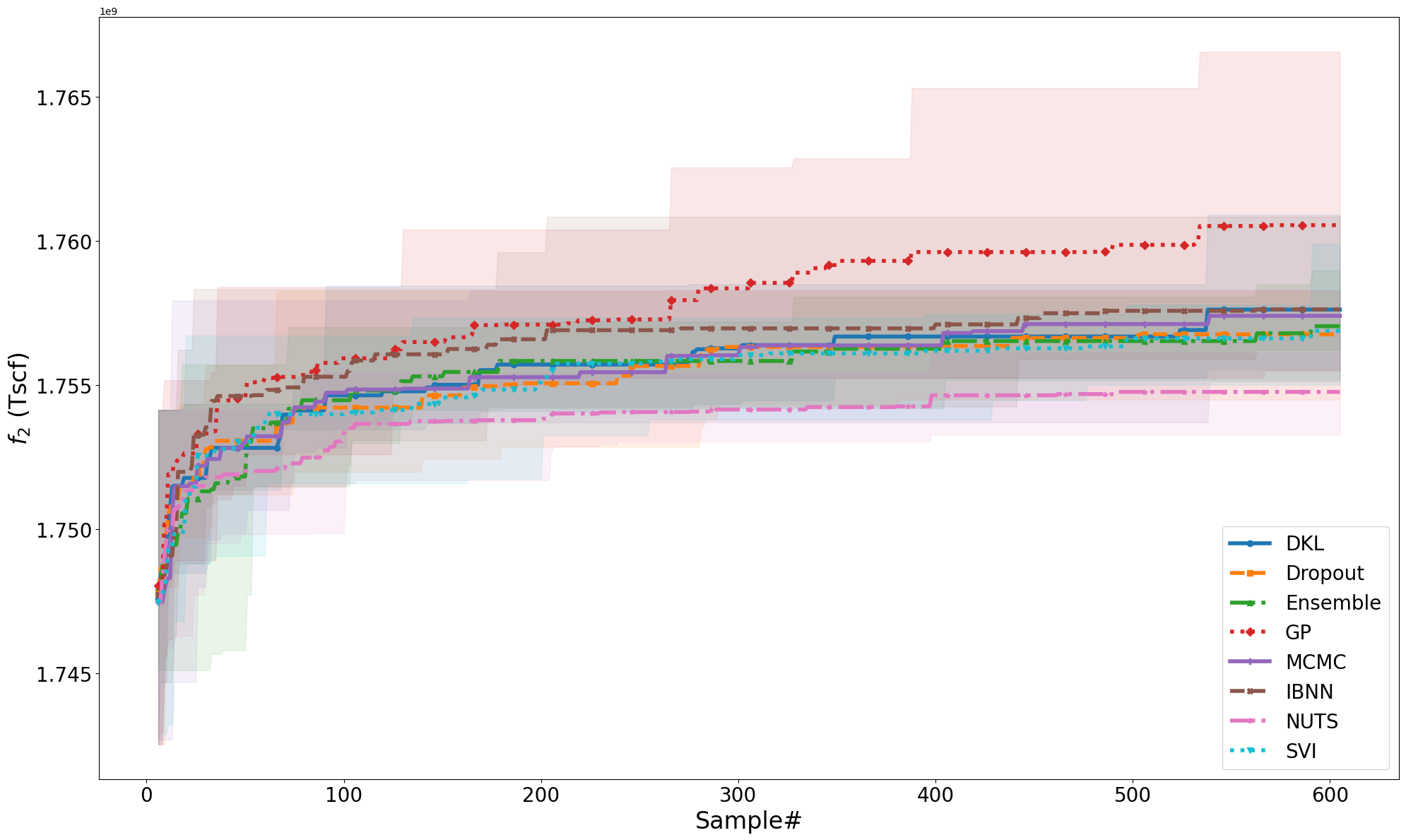}
\caption{Supplementary convergence of the maximum second objective value vs iteration number\label{fig12}}
\end{figure}

\begin{figure}[H]
\includegraphics[width=0.7\linewidth,keepaspectratio]{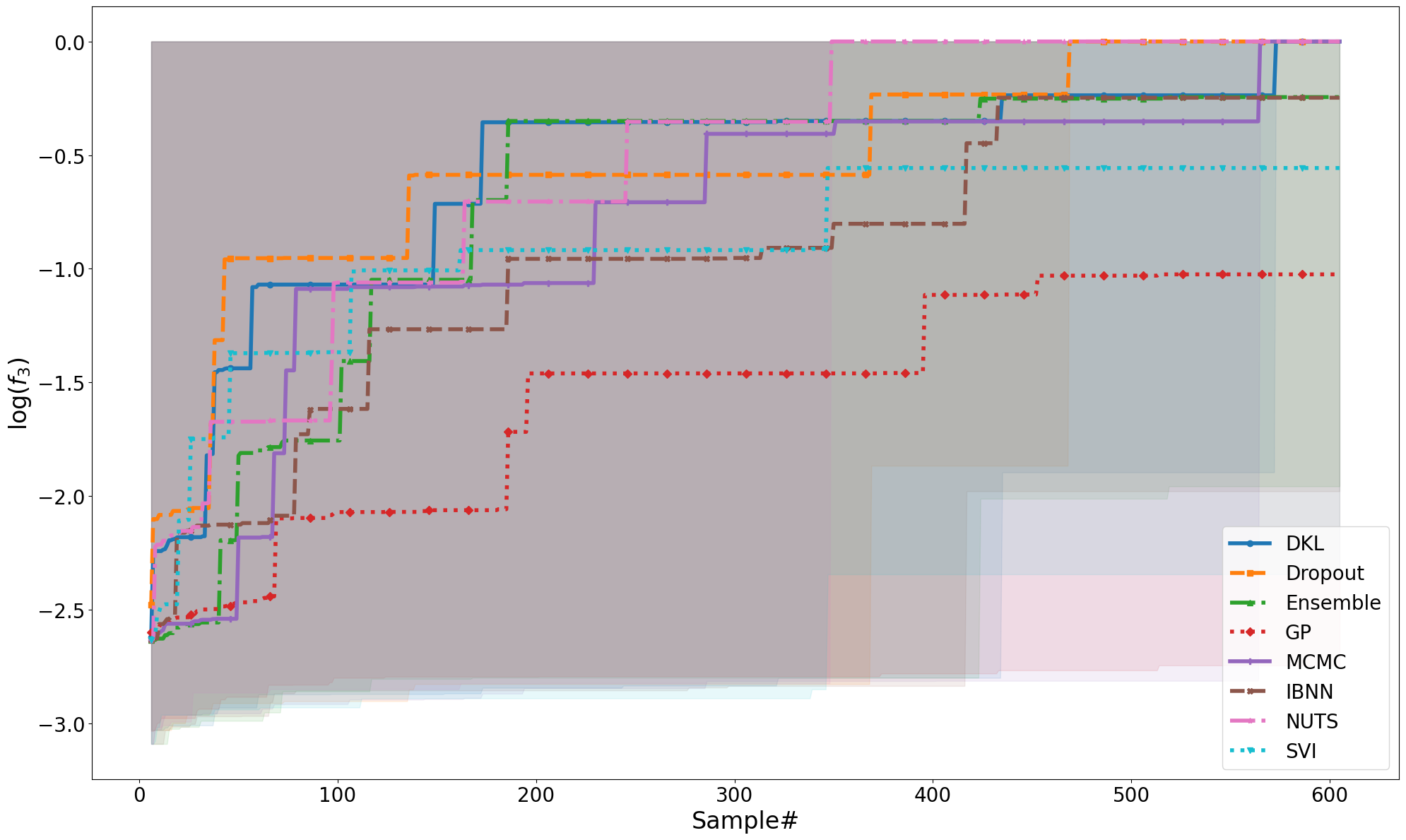}
\caption{Supplementary convergence of the maximum third objective value vs iteration number\label{fig13}}
\end{figure}

\begin{figure}[H]
\includegraphics[width=0.7\linewidth,keepaspectratio]{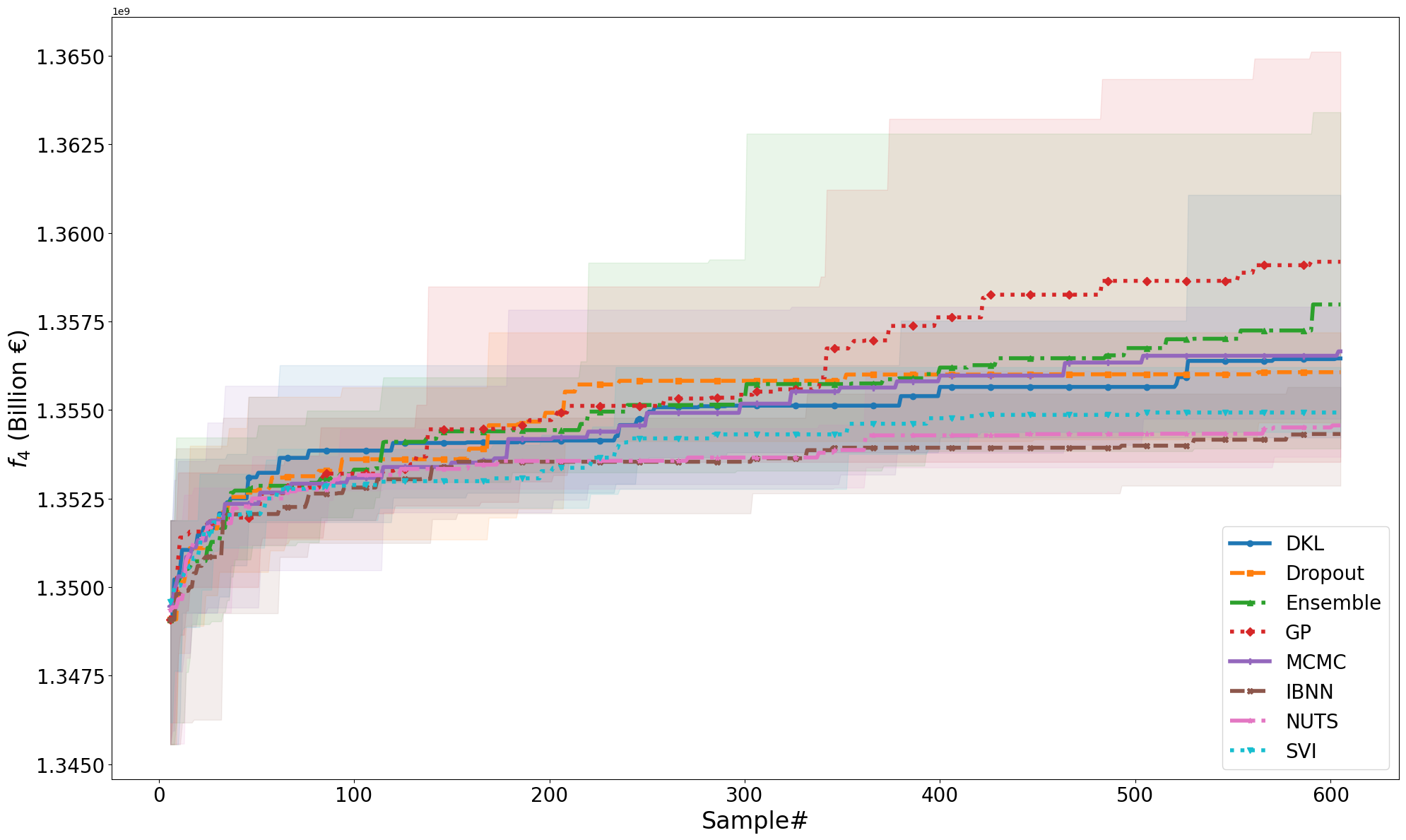}
\caption{Supplementary convergence of the maximum fourth objective value vs iteration number\label{fig14}}
\end{figure}

For completeness, Figures \ref{fig11}--\ref{fig14} show the evolution of the best objective-wise values found by each model across objective evaluations. Since each model was tested across eight trials, dashed lines indicate the average trajectory and the shaded regions indicate the range between minimum and maximum values across trials.

In Figure \ref{fig11}, the evaluation results for $f_1$ are presented. While most models converge on average to a similar optimal value, the GP model exhibits significantly lower average values. This discrepancy can be partially explained by the shaded region, which represents the range of values encountered across trials. The broad spread for GP suggests that in at least one trial, the model never approached the optimal value of the first objective, leading to consistently lower performance. On average and on the maximum case, the SVI model found the most optimal values for the first objective, followed by IBNN and Dropout.

In Figure \ref{fig12}, the results for the total sequestration volume $f_2$ are presented. This is one of the most critical objective functions, as it is commonly used to assess the success of a CCS project in its most simplified form, without considering economic, contractual, or safety constraints. Most models demonstrate a sharp initial improvement in their predictions, followed by a more gradual refinement over time. On average, the performance of most models is comparable; however, the GP model achieves the highest average results, consistently outperforming the others in maximizing total sequestration. In contrast, the NUTS model exhibits the lowest average performance, with a noticeably slower convergence towards higher sequestration volumes. The spread of shaded regions further indicates that the variability across trials is relatively limited for most models, suggesting stable performance trends.

The penalty function $f_3$ results are presented in Figure \ref{fig13} in logarithmic scale. The wide shaded areas indicate that, in at least one trial, each model reached the maximum possible $f_3$ value. Notably, the NUTS, MCMC and DKL models consistently found such solutions across all trials, whereas the GP model exhibited the lowest overall average performance. The presence of shaded regions near the value of $-3$ suggests that in many trials, some models never explored solutions that maximize the third objective. In general, values greater than $-2$ for this objective correspond to unconventional solutions characterized by high recycling rates. This explains the large variability observed in the shaded regions of the plot, reflecting the distinct exploration strategies adopted by different models across various trials.

Results for the $f_4$ (NPV) value are shown in Figure \ref{fig14}. This is a key objective in a CCS project, as it ultimately determines whether stakeholders choose to proceed with the operation. The NPV used in this analysis is simplified, as drilling costs were omitted and operational pricing was based on speculative estimates. However, this simplified NPV is sufficient to act as a counterweight to extreme production/recycling under the present control parameterization. Once again, the GP model demonstrates the best average performance, followed by the Ensemble model. The convergence pattern across all models follows a typical trend: a rapid initial increase, followed by a more gradual improvement in later iterations. However, this gradual increase appears more pronounced compared to the results in Figure \ref{fig12}, which corresponds to the second objective function (total sequestration). This suggests that optimizing for NPV requires a more extended search process before stabilizing at near-optimal values.

\section{Additional ablation study: GP kernel choice on Case Study 1}
\label{app:kernel_ablation_case1}

A sensitivity study is performed on Case Study 1 (Variation 1) by comparing the SE kernel  with the Mat\'ern one which is another popular choice. The purpose of this experiment is to assess whether the main conclusions are materially affected by the GP kernel choice.

The experimental setup was kept identical to the one used in the main study for this case: same optimization problem, same BO loop, same initialization strategy and the same evaluation budget. Only the GP kernel was changed. We repeated the experiment over multiple independent trials and report the best-so-far objective value as a function of the BO iteration, showing the mean together with one standard deviation across trials.

\begin{figure}[h]
	\centering
	\includegraphics[width=0.82\textwidth]{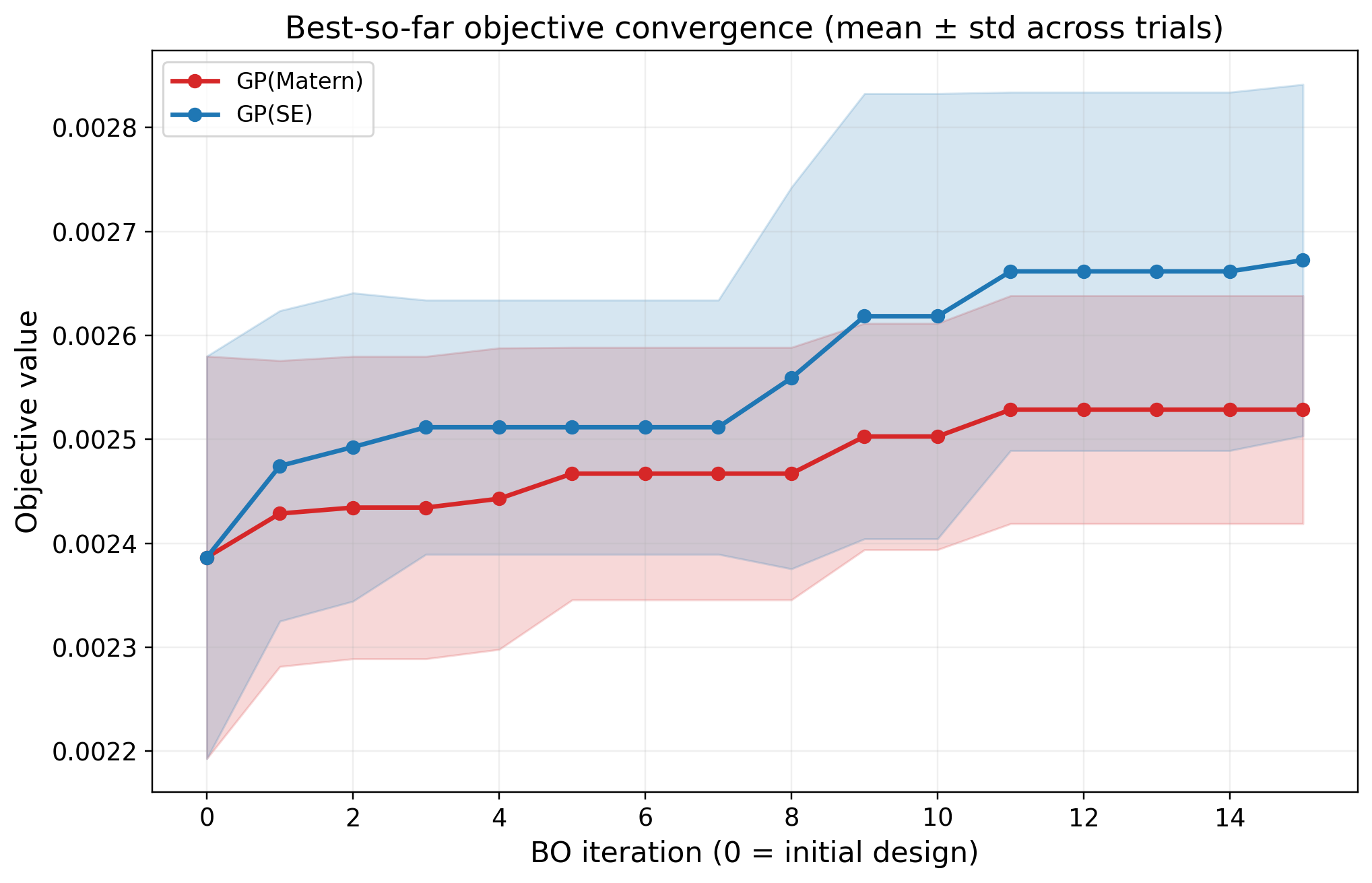}
	\caption{Best-so-far objective convergence for the GP baseline on Case Study~1 (Variation~1), comparing the Mat\'ern and squared exponential (SE) kernels. Solid lines show the mean across trials and shaded regions indicate $\pm 1$ standard deviation.}
	\label{fig:appendix_gp_kernel_ablation_case1}
\end{figure}

Figure~\ref{fig:appendix_gp_kernel_ablation_case1} shows the resulting convergence curves. The SE kernel achieves higher best-so-far objective values on average over most of the optimization budget, while the Mat\'ern kernel remains competitive and follows a similar convergence pattern. Importantly, the two curves are of the same order and their uncertainty bands overlap for a substantial part of the optimization trajectory. 

\newpage


\bibliographystyle{elsarticle-num} 
\bibliography{references}

\end{document}